\documentclass[journal]{IEEEtran}
\usepackage{graphics}
\usepackage{epsfig}
\usepackage[utf8]{inputenc}
\usepackage[english]{babel}
\usepackage{nomencl}
\usepackage{threeparttable}
\makenomenclature
\usepackage{times} 
\usepackage{graphicx}
\usepackage{float}
\usepackage{algorithm}
\usepackage{algorithmic}
\usepackage[algo2e,ruled]{algorithm2e}
\usepackage{tikz}
\usetikzlibrary{shapes,arrows}
\usepackage{verbatim}
\usepackage{amsmath} 
\usepackage{amssymb}  
\usepackage{amsmath}
\usepackage{amssymb}
\usepackage{mathtools}
\usepackage{mathrsfs}
\usepackage{graphicx}              
\usepackage{multirow}
\usepackage{amsthm}
\usepackage{booktabs}
\graphicspath{{TMech2018Figs/}}

\title{A deep learning-based remaining useful life prediction approach for bearings}
\author{Cheng Cheng, Guijun Ma, Yong Zhang, Mingyang Sun, Fei Teng, Han Ding, and Ye Yuan
\thanks{This work was supported by the National Natural Science Foundation of China [Grant number 91748112] and by the Primary Research \& Development Plan of Jiangsu Province [Grant number BE2017002].}
\thanks{C. Cheng and Y. Yuan are with the Key Laboratory of Image Processing and Intelligent Control, School of Artificial Intelligence and Automation,  Huazhong University of Science and Technology, Wuhan, 430074, China. }
\thanks{Y. Yuan, G. Ma and H. Ding are with the State Key Laboratory of Digital Manufacturing Equipment and Technology, Huazhong University of Science and Technology, Wuhan 430074, China.}
\thanks{G. Ma and H. Ding are with the School of Mechanical Science and Engineering, Huazhong University of Science and Technology, Wuhan 430074, China.}
\thanks{Y. Zhang is with School of Information Science and Engineering, Wuhan University of Science and Technology, Wuhan 430081, China.}
\thanks{M. Sun is with the College of Control Science and Engineering, Zhejiang University, Hangzhou, 310007, China.}
\thanks{F. Teng is with the Department of Electrical \& Electronic Engineering, Imperial College London, London, SW7 2AZ, UK.}
\thanks{For correspondence, contact Prof. Ye Yuan (yye@hust.edu.cn).}
}

\begin{document}

\maketitle

\begin{abstract}
In industrial applications, nearly half the failures of motors are caused by the degradation of rolling element bearings (REBs). Therefore, accurately estimating the remaining useful life (RUL) for REBs is of crucial importance to ensure the reliability and safety of mechanical systems. To tackle this challenge, model-based approaches are limited by the complexity of mathematical modeling. Conventional data-driven approaches, on the other hand, require massive efforts to extract the degradation features and construct the health index. In this paper, a novel data-driven framework is proposed to exploit the adoption of deep convolutional neural networks (CNN) in predicting the RULs of bearings. More concretely, raw vibrations of training bearings are first processed using the Hilbert-Huang transform to construct a novel nonlinear degradation energy indicator which can be used as the training label. The CNN is then employed to identify the hidden pattern between the extracted degradation energy indicator and the raw vibrations of training bearings, which makes it possible to estimate the degradation of the test bearings automatically. Finally, testing bearings' RULs are predicted through using a $\epsilon$-support vector regression model. The superior performance of the proposed RUL estimation framework, compared with the state-of-the-art approaches,  is demonstrated through the experimental results. The generality of the proposed CNN model is also validated by performance test on other bearings undergoing different operating conditions.
\end{abstract}


\vspace{-2em}
\section*{Nomenclature}
\addcontentsline{toc}{section}{Nomenclature}
\begin{IEEEdescription}[\IEEEusemathlabelsep\IEEEsetlabelwidth{FT or $L_{ft}$}]
\item[CNN] Convolutional neural network.
\item[DEI] Degradation energy indicator.
\item[EMD] Empirical mode decomposition.
\item[ETA] Exponential transformed accuracy.
\item[HHT] Hilbert-huang transform.
\item[IMF] Intrinsic mode function.
\item[MAE] Mean average error.
\item[NRMSE] Normalized root mean square error.
\item[MSE] Mean square error.
\item[REB] Rolling element bearing.
\item[RUL] Remaining useful life.
\item[SVR] Support vector regression.
\item[FT or $L_{ft}$] Failure threshold.
\item[$N$] Length of historical units of training bearing.
\item[$Q$] Length of historical units of test bearing.
\item[$U$] No. of predicted units of test bearing.
\item[$\bf{S_i}$] Sensor measurement signal of $i$-th unit.
\item[$\Delta t$] Sampling time.
\item[$P$] Number of measurements in $i$-th unit.
\item[$\tau$] Time interval between two recording phases.
\item[$\bf{L}$] DEI of training bearing.
\item[$\bf{L_{norm}}$] Normalized DEI of training bearing.
\item[$\bf{L_{test}}$] Estimated DEI of test bearing.
\item[$\bf{\widehat{L}_{U,test}}$] Predicted DEI of test bearing.
\item[$K$] Total number of layers.
\item[$W^k$] Weights in $k$-th convolutional layer.
\item[$B^k$] Bias in $k$-th convolutional layer.
\item[$\widehat{T}_{failure}$] Predicted RUL of test bearing.
\item[$T_{failure}$] Real RUL of test bearing.
\item[$\beta$] Number of test bearings.
\item[$E_r\%$] Relative percentage error of prediction.
\item[$S_{mean}$] Average score of prediction.
\item[$\bf{X}$] Training set for SVR.
\item[$\mathbb{R}$] The set of real numbers.
\item[$\mathbb{Z}$] The set of positive intergers.
\end{IEEEdescription}

\section{Introduction}
\label{sec:Intro}

\IEEEPARstart{T}{o} constrain relative motion while reducing friction between moving parts, rolling element bearings (REBs) are one of the most widely used elements in industrial machinery. Prognostics and health management of bearings is of significance for safety, reliability and effectiveness of the mechanical systems \cite{harris2001rolling,yuan2017bayesian}. The literature \cite{heng2009rotating} shows nearly half of motor failures are related to the degradation of bearings. As such, estimating the remaining useful life (RUL) (i.e., time-to-failure prognostics) of bearings has attracted a great deal of attention in recent years \cite{rodrigues2018remaining}. RUL prediction helps users monitor the condition of the bearings and provides an estimation of time left before a failure occurs.  Compared with fault diagnosis, which has been well investigated over past few decades \cite{de2016data}, the problem of RUL prediction studied in this paper is a relatively new and challenging topic due to the huge amount of uncertainties of environment and operating condition.

In general, RUL prediction approaches can be categorized into model-based and data-driven approaches. Model-based approaches aim to build a physical model to represent the degradation of the rolling bearing \cite{wang2018real}. Li \textit{et al.} \cite{li2000stochastic} predicted the defect growth on a bearing unit using Paris's law for fatigue. However, it is difficult to construct a precise physical degradation model due to the sensitivity of the model parameters and noised operating environments. This limits the practical applications of the model-based approaches. On the other hand, data-driven approaches benefit from the extensive expertise in signal processing and machine learning \cite{si2011remaining}, and infer the degradation process of bearings without knowing any physics of degradation failure. The prognostic framework of the data-driven approaches mainly consists of three stages: 1) feature extraction from noisy sensory signals, which helps to build up the health indicator for the learning of system degradation behavior; 2) degradation models are trained on the training bearing using statistical or machine learning techniques; and 3) the degradation indicator of the test bearings can be estimated based on the model trained in the second stage. Then, the unknown degradation process can be predicted by applying regression techniques (i.e., $\epsilon$-SVR).

To extract features from raw signals, time-domain, frequency-domain, and time-frequency domain analysis are commonly adopted. Among them, the time-frequency analysis has been found to be the most efficient due to its ability to characterize transient signals over time and frequency domains \cite{wang2019multi}. Well-known time-frequency techniques for extracting bearing features include short-time Fourier transform \cite{gao2015feature}, wavelets \cite{yan2014wavelets}, Wigner-Ville distribution \cite{meng1991rotating}, and Hilbert-Huang transform (HHT) \cite{soualhi2015bearing}. Implementation of the short-time Fourier transform is limited by its time-frequency resolution capability; for instance, low frequencies are difficult to identify with short windows. On the other hand, wavelets and the Winger-Ville distribution provide richer pictures than short-time Fourier transform; however, their effectiveness depend on estimating the Hurst parameter and the quality of the analyzed signal. HHT shows better computational efficiency and resolution over other time-frequency analysis. which uses the techniques of empirical mode decomposition (EMD) and Hilbert transform (HT) to decompose the original vibration signal into a number of intrinsic mode functions (IMFs) in various frequency scales. Frequency components of each IMF are related to both the sampling frequency and the signal itself, thus demonstrating that HHT is a self-adaptive signal processing technique perfectly suited to non-stationary signals. Wu \textit{et al.} \cite{wu2018degradation} analyzed the time-to-failure prognostics of REBs, which extracts ten statistical features using time and frequency analysis and eleven IMF features using HHT time-frequency analysis. The gear fault identification method proposed in \cite{cheng2013gear} is based on the HHT, and first six IMFs are selected as inputs for SOM neural networks for fault diagnosis.

Data-driven RUL prediction approaches are mainly based on statistical and machine learning techniques, such as artificial neural networks (ANN) \cite{tian2012artificial}, fuzzy logic systems \cite{wang2007adaptive}, and auto-regressive (AR) models \cite{sikorska2011prognostic}. The computational cost of ANN is relatively high in terms of optimizing the weights of the model. The performances of the AR models and fuzzy logic systems require precise trend of historical observations and high-quality training data, respectively. Recently, deep learning has merged into research and industry fields, and has beaten other machine learning techniques in speech recognition and image recognition tasks \cite{hinton2012deep}. Deep learning model is good at discovering high level abstractions from labeled data using a back-propagation algorithm \cite{sun2019deep}. Specifically, it learns feature representations automatically rather than designing the hand-created features by experience. As the most well-known model in deep learning, in recent years, CNN dominates the recognition and detection problems in computer vision domain, which is distinguished by three characteristics, namely local connections, shared weights, and local pooling \cite{lecun2015deep,yuan2019nsr}. The first two characteristics show that the CNN model requires fewer parameters to detect local information than multilayer perceptron, while the last characteristic ensures shift invariance to the networks. Typically, 1-D CNN will be employed to this work to learn the latent space of input sensory time-series vibrations, which has been applied with great success to speech recognition and document reading tasks. Few attempts have been made for the prediction problem using CNN-based models \cite{chen2019real, zhu2018estimation, ren2018prediction, sun2020using}. This paper exploits the adoption of CNN technique in estimating the RUL of bearings, as a prognosis problem, to learn about the nonlinear degradation behavior according to raw vibration data and an extracted label. Instead of using the CNN technique to perform the time-series prediction, the main function of the CNN model in this paper is to reveal the hidden dependencies between the vibration data and the DEI of the training bearing, which makes full use of the advantages of CNN in automatic feature extraction.  

In this work, we propose a data-driven framework for predicting the RUL of REBs by applying the HHT, CNN, and $\epsilon$-Support vector regressorion ($\epsilon$-SVR). The raw vibration signals collected from sensors are processed by the HHT method and a novel time-series degradation indicator, i.e., DEI, is constructed. Subsequently, a CNN model is trained to learn the features from the input raw vibration to the DEI label on the training bearings, and used to predict the DEIs of testing bearings. Then, a $\epsilon$-SVR model is introduced so that the evolution of the degradation can be forecast till the bearing failure. The effectiveness of the proposed based framework for RUL prediction is validated on an experimental platform (i.e., PRONOSTIA). Much lower RUL prediction errors are achieved, compared with eight existing approaches in previous papers and two tested methods designed in this paper, indicating the superior performance of the proposed method.

\begin{figure}[htb!]
              \centering
              \includegraphics[width=0.8\columnwidth]{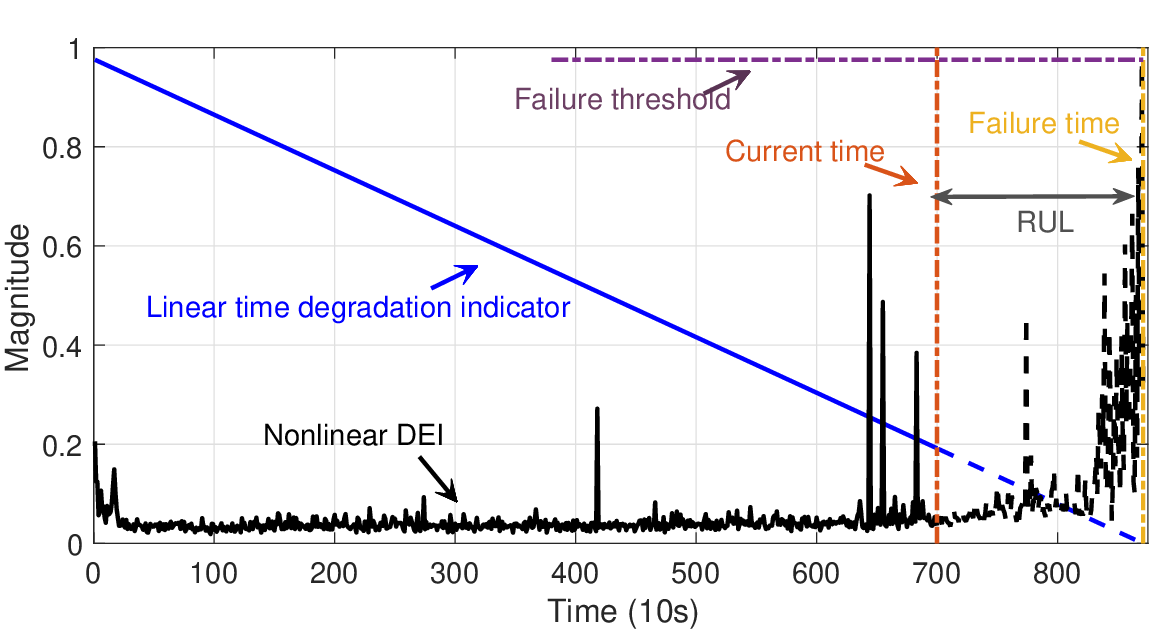}\\
              \caption{Linear time degradation indicator (blue) vs. nonlinear DEI (black), as a label for network training. The nonlinear DEI experiences a long time flat curve before a sharp degradation trend when it gets close to the end of bearing lifetime, which reflects the real degradation process of most machinery systems \cite{lei2018machinery}. With the aggregation of damages at different bearing components close to the end of lifetime, the simple time degradation indicator is less effective than DEI for RUL estimation.}
                \label{fig:HIcomp}
  \end{figure}

This work makes the following contributions: 1) The proposed method successfully extracts a novel nonlinear degradation energy indicator (DEI) (see Fig. \ref{fig:HIcomp}, compared with the linear time degradation indicator) to describe the degradation trend of the training bearing, according to the nature frequencies of bearing components; 2) The proposed CNN architecture is general and robust for similar operation conditions, it can transfer to another bearing undergoing different operating condition and obtain good prediction results, without changing CNN hyper parameters and the depth of layers; 3) The propose DEI is an integrated indicator with regards to the maximum vibration levels among different bearing components, which considers all the possible detects on the rolling element bearing. This is a more realistic indicator as the localized defects are not initially initiated in real industrial applications, meaning that all the types of defects have to be considered; and 4) CNN scales all the indicators of training bearings and test bearings into a consistent latent space. Thus, training and testing can share a same failure threshold (FT), i.e., the maximum value of the indicator for the training bearing.

The outline of this paper is as follow. Section \ref{sec:Main_RUL} presents the proposed RUL prediction framework with technical details. In Section \ref{sec:tests}, experimental results obtained from bearing degradation tests are carried out. Thereby, the performance of the proposed framework is validated and the results show improved accuracy in predicting the RUL compared with eight state-of-the-art approaches and two designed test methods in this paper. Section \ref{sec:conclusion} summarizes the paper and discusses future works.

\vspace{-0.5em}
\section{Degradation indicator training and RUL prediction algorithm}
\label{sec:Main_RUL}
\begin{figure*}[!htb]
\centering
\includegraphics[width=1.5\columnwidth]{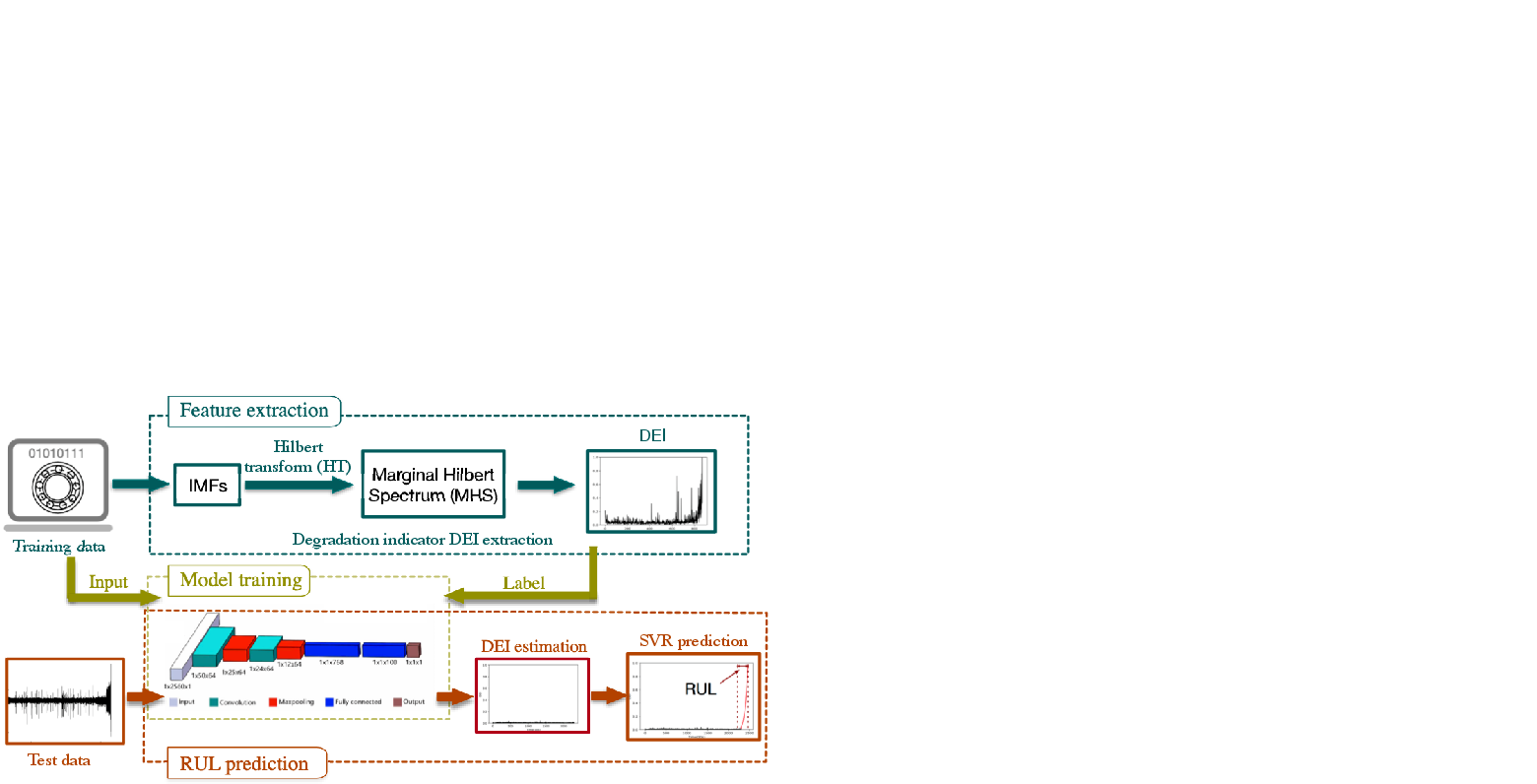}\vspace{-0.5em}
\caption{Data-driven RUL prediction framework. This framework involves three stages: (1) a feature extraction stage aims to extract the degradation indicator. The DEI is extracted using EMD, HT and marginal Hilbert spectrum (MHS). The value of DEI is also related to the nature frequencies of different bearing components; 2) a model training stage aims to train a CNN model based on the training data that discovers the hidden pattern between the DEI from and the raw vibration data; and (3) a prediction stage predicts the RUL according to the trained CNN model and a $\epsilon$-SVR forecasting model.}
\label{fig:platform}
\end{figure*}

The overall framework for the prediction of the RUL can be decomposed into three parts. The schematic of the overall framework is shown in Fig. \ref{fig:platform}. The key challenges of this work involve: obtain the DEI to represent the degradation behaviour; establish a CNN model to map raw vibration signal to the DEI; and construct an $\epsilon$-SVR to predict the RUL. Thus, in the following subsections, the explicit expression of degradation feature extraction, CNN model, and $\epsilon$-SVR forecasting model will be derived in Section \ref{sec:HHT}, Section \ref{sec:CNN}, and Section \ref{sec:SVR}.

\subsection{Degradation indicator extraction}
\label{sec:HHT}
To begin with, for a training bearing, it is assumed that the raw vibration signal till the end of lifetime with $N\in \mathbb{Z}$ historical units have been acquired. The sensory vibration signal $\mathbf{S_i}=\{S_{i}(t_{p})\}^P_{p=1}$, with sampling time $\Delta t$, is measured at each historical unit for $i\in D=\left \{ 1,2,..., N\right \}$, where $P$ is the number of measurements recorded in each historical unit.

EMD is a self-adaptive method which is normally applied to analyze non-stationary and nonlinear signals. It decomposes the raw vibration data $\bf{S_i}$ into $n$ number of IMFs, illustrating the natural oscillation modes from fast to low oscillations. 

For the $i$-th unit, the $j$-th mode ($j=1,\ldots,n$) of IMF, IMF$_{i,j}(t_p)=a^v_{i,j}(t_p)$ for $p=1,\ldots, P$, is calculated iteratively associated with the iteration number $v=0,1,\ldots$. 

First let $v=0$ and initialize $a^0_{i,j}(t_p)$ for $p=1,\ldots, P$ by:
\begin{equation}
a^0_{i,j}(t_p)= \left\{\begin{matrix}{S_i}(t_p), \,\,\,&j=1\\ {S_i}(t_p)- \sum_{j=1}^{n-1}\textup{IMF}_{i,j}(t_p), \,\,\,&j=2,..., n \end{matrix}\right. .
\label{eq:IMF2}
\end{equation}

Define IMF$_{i,j}(t_p)=a^0_{i,j}(t_p)$ for $p=1,\ldots, P$ only if the IMF meets the following two conditions:
\begin{description}
\item[(I)] The IMF should have one or zero difference between the extrema number and the number of zero crossing,
\item[(II)] along the time axis, the average value of upper and lower bound of the IMF should be zero everywhere.
\end{description}

Otherwise, update the IMF$_{i,j}(t_p)=a^v_{i,j}(t_p)$ for $p=1,\ldots, P$ through the following iteration procedure with $v=1,2,\ldots$ until the IMF satisfies both aforementioned conditions (I) and (II):
\begin{equation}
a^v_{i,j}(t_p)=a^{v-1}_{i,j}(t_p)-\omega^{v-1}_{i,j}(t_p), \,\,\forall p,
\label{eq:IMF1}
\end{equation}
 where $\omega^{v-1}_{i,j}(t_1),\ldots,\omega^{v-1}_{i,j}(t_P)$ are the mean values of the upper envelope and the lower envelope of $a^{v-1}_{i,j}(t_1),\ldots,a^{v-1}_{i,j}(t_P)$. 

Once obtaining all IMFs, the analytical form of each IMF can be written as: 
\begin{equation}
\textup{IMF}^A_{i,j}(t_p)=\textup{IMF}_{i,j}(t_p)+\textit{\textbf{i}}\,\textup{IMF}^H_{i,j}(t_p).
\end{equation}
where \textit{\textbf{i}} is the imaginary part of the $\textup{IMF}^A_{i,j}(t_p)$. $\textup{IMF}^H_{i,j}(t_p)$ is the Hilbert transformation of $\textup{IMF}_{i,j}(t_p)$ by convolution with function $\frac{1}{\pi t}$, given as:
\begin{equation}
\textup{IMF}^H_{i,j}(t_p)=\frac{1}{\pi}\int_{-\infty}^{+\infty} \frac{\textup{IMF}_{i,j}(s)}{t_p-s}ds.
\end{equation}

By this means, we can calculate the instantaneous amplitude $h_{i,j}(t_p)$ and phase $\phi_{i,j}(t_p)$
\begin{equation}
\begin{array}{l}
h_{i,j}(t_p)=\sqrt{(\textup{IMF}_{i,j}(t_p))^2+(\textup{IMF}_{i,j}^{^H}(t_p))^2} , \\
\phi_{i,j}(t_p)=\arctan\left(\frac{\textup{IMF}^H_{i,j}(t_p)}{\textup{IMF}_{i,j}(t_p)}\right).
\end{array}
\end{equation}

Accordingly, it is easy to derive the instantaneous frequency  
\begin{equation}
f_{i,j}(t_p)=\frac{1}{2\pi}\frac{\phi_{i,j}(t_p+\Delta t)-\phi_{i,j}(t_p-\Delta t)}{2\Delta t}.
\end{equation}
for $p=1,\ldots,P$.

Then, the Hilbert Spectrum of $\bf{S_i}$ is obtained by 
\begin{equation}
M_{i}(f_i,t_p)= \sum_{j=1}^{n} h_{i,j}(f_{i,j},t_p).
\end{equation}
The marginal Hilbert spectrum (MHS) $M_{i}(f_i)$ can be written as: 
\begin{equation}
M_{i}(f_i)=\sum_{p=1}^P M_{i}(f_i,t_p).
\end{equation}

Nature frequencies of bearing components depend on the geometry of the bearing and its rotation speed. Expression of these frequencies are given in Table \ref{tb:bearing_f}, in where $\eta$ is the number of balls, $f_{\omega}$ is the rotation frequency, $\Phi$ is the contact angle, and $\ell_{ball}$ and $\ell_{pitch}$ are the ball diameter and pitch diameter, respectively.

\begin{table}[!]
\centering
\caption{Bearing frequencies of inner race, outer race and ball\cite{tandon1999review}}
\label{tb:bearing_f}
\begin{tabular}{lll}
\toprule
Symbol & Description          & Expression \\ \hline
\midrule
$f_{inner}$  & Inner race frequency &    $\frac{\eta}{2}\cdot f_{\omega}\cdot \left ( 1+\frac{\ell_{ball}}{\ell_{pitch}}\cdot \cos\Phi \right )$        \\ 
$f_{outer}$  & Outer race frequency &    $\frac{\eta}{2}\cdot f_{\omega}\cdot \left ( 1-\frac{\ell_{ball}}{\ell_{pitch}}\cdot \cos\Phi \right )$          \\ 
$f_{ball}$  & Ball frequency       &    $\frac{\ell_{ball}}{\ell_{pitch}}\cdot f_{\omega}\cdot \left ( 1-\frac{\ell^2_{ball}}{\ell^2_{pitch}}\cdot \cos^2\Phi \right )$          \\ \bottomrule
\end{tabular}
\end{table}

With the bearing frequencies of different components, the value of the DEI, $L_i$, at historical unit $i$ is defined as the maximum value of the MHS by substituting the $f_{inner}$, $f_{outer}$ and $f_{ball}$ into $M_{i}(f_i)$, given that 
\begin{equation}
L_i = {\max\limits_{f_i\in\{f_{inner},f_{outer},f_{ball}\}}}M_i(f_i), \,\,\,i\in D.
\label{eq:mhs}
\end{equation}

The extracted DEI $\bf{L}$$=[L_{1},...,L_N]$ is nromalized before training:
                \begin{equation}
                L_{norm,i}=\frac{L_i-\min(\bf{L})}{\max(\bf{L})-\min(\bf{L})}\pm{\epsilon},    \epsilon \to 0
                \end{equation}
for $i=1,\ldots,N$, and $\epsilon$ is an infinitesimal that used to avoid the value of label equal to 0 or 1. Thus, the normalized DEI is 
\begin{equation}
\mathbf{L_{norm}}=[L_{norm,1},...,L_{norm,N}].
\end{equation}

\subsection{DEI pattern learning}
\label{sec:CNN}
In this work, layers with repeated components are stacked in a CNN architecture, including convolutional layers, pooling layers, fully connected layers, and a regression layer \cite{lecun2015deep}.
 \begin{figure*}[htb!]
 \centering
\includegraphics[width=1.8\columnwidth]{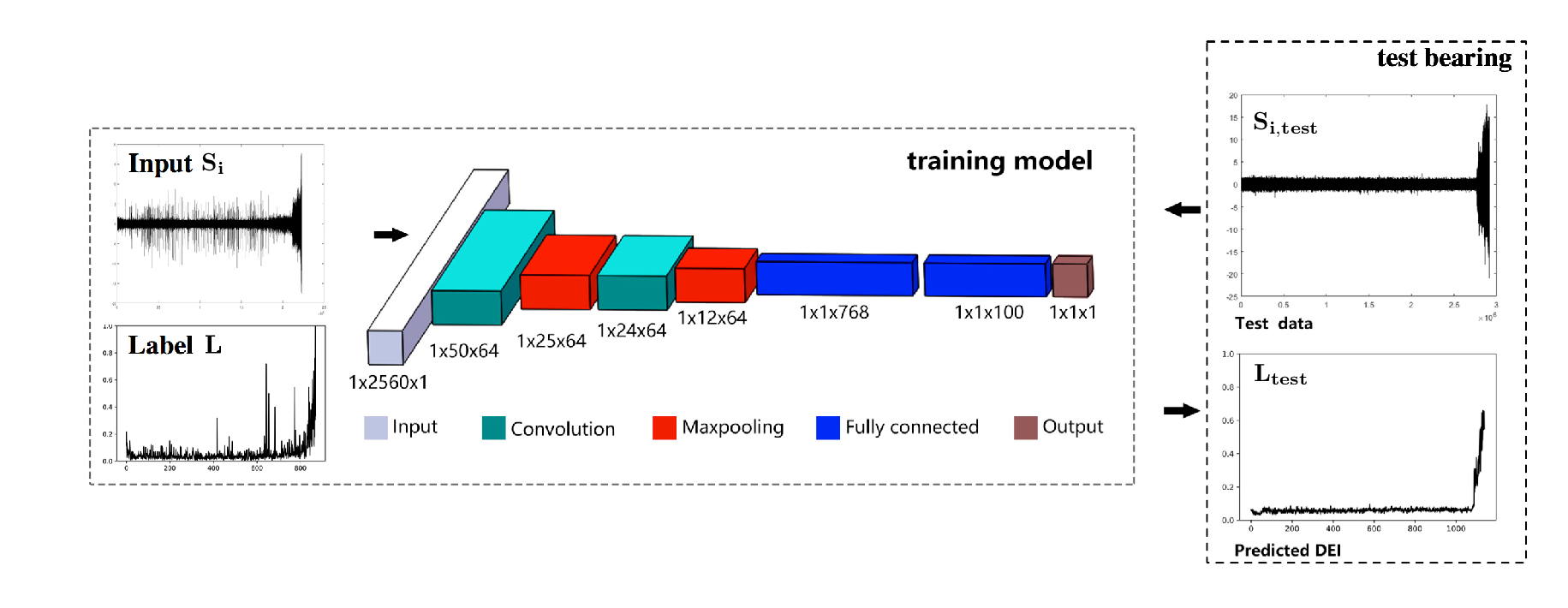} \vspace{-1em}
\caption{The proposed CNN architecture. Normalized DEI $\bf{L}$ is used as the label for training. A 6-layer CNN model is trained to map the raw vibration data $\bf{S_{i}}$ (input) to the DEI $\bf{L}$ (output). For a new test bearing, the vibration data $\bf{S_{i,test}}$ is directly input to the CNN model to obtain the estimated DEI $\bf{L_{test}}$.}
\label{fig:cnnmodel}
\end{figure*}

\textit{Convolutional layer} contains organized patches in convolutional layers, each patch is calculated by composing the features of the previous layer through a filter bank with the following equation:
\begin{equation}
u^k_m=W^k\ast u^{k-1}_e+B^k,
\label{eq:sigment_conv}
\end{equation}
where $u^k_m\in \mathbb{R}$ denotes the output of the $m$-th unit in the $k$-th layer.  $u^{(k-1)}_e\in \mathbb{R}^{1\times o^{k}}$ is the input data the $e$-th sub-vector in the previous layer $k-1$, where $o^{k}$ is kernel size in layer $k$. $W^k\in \mathbb{R}^{1\times o^k}$ and $B^k\in \mathbb{R}$ denote the connecting weights and bias in the $k$-th layer, respectively. `$\ast$' means the convolution operation. It is noted that when $k=1$, $u^{(k-1)}_e$ is a sub-vector of the raw vibration data $\bf{S_i}$. Here we define all neurons in each layer is ${\bf{u}}^k=[u^k_1,\ldots,u^k_m,\ldots,u^k_{G^k}]$ for $k\in D_k=$ \{1,2,$\ldots$,$K$\}, where $G^k\in \mathbb{Z}$ is the number of neurons in the $k$-th layer and $K$ is the total number of layers. For convolutional layer, $G^k=(G^{k-1}-o^{k})/I^k_{cv}+1$ and $I^k_{cv}\in \mathbb{Z}$ is the stride in convolutional layer.

\textit{Activation function} is introduced after convolutional layer. Among various activation functions, Rectified linear unit (ReLU) ${r}^k_m=\max(0,{u}^k_m)$ is chosen as the nonlinear activation function to prevent the issue of vanishing gradient which may significantly increase the training time or even lead to the non-convergence. 

\textit{Pooling layer} is then used as a nonlinear down-sampling layer to extract the maximum feature values in each patches of the input data. Its function is to save computation time and downsize the number of parameters of the model as well as control overfitting. More specifically, pooling transforms small windows into single values by maxing or averaging. Consequently, the features extracted within the small window are similar and therefore illustrating the shift invariance property of CNN. Max-pooling layer is selected in this work as it is an algorithmic choice to ensure the generalization of neural networks \cite{neyshabur2017exploring}, which is given by:
\begin{equation}
           P^k_m =  \max^{}_{\gamma=1,\ldots,\lambda^k}\,\,r^{(k-1)}_{\gamma+(m-1)I^k_{pl}},
\label{eq:sigment_ply}
\end{equation}
where $\lambda^k\in \mathbb{Z}$ is the pooling size, and $I^k_{pl}$ is the stride in max pooling layer.

\textit{Fully connected layer and regression layer}, like a classic ANN network, take the results of the convolution and max-pooling processes and use them to generate a predicted label. Since we use a normalized DEI $\mathbf{L_{norm}}\in \mathbb{R}^{1\times N}$ as the label for learning,  the sigmoid function $\text{sigm}({\bf{u}}^{K-1})$ with an output value between (0,1) is applied to the last layer for normalized output. Hereby, mean square error (MSE) function is used to compute the loss with the expression:
\begin{equation}
z =  \frac{1}{N}\sum_{i=1}^{N}(L_{norm,i}-\tilde{L}_{norm,i})^2,
\label{eq:lossfx}
\end{equation}
in where the proposed CNN model is minimizing the loss function $z$ between ground label DEI $\mathbf{L_{norm}}$ and predicted label $\tilde{\bf{L}}_{\bf{norm}}$. Algorithm 1 outlines the proposed CNN modeling procudure.

\SetKwRepeat{Do}{do}{end}%
\begin{algorithm}
\label{alg:FramworkCNN}
    \SetKwInOut{Input}{Input}
    \SetKwInOut{Output}{Output}

    \Input{The normalized extracted DEI label $\bf{L}_{norm}$;\\ The raw vibration data $\{\bf{S_i}\}^N_{i=1}$.\\}

    \Output{Trained CNN parameters: $W^k$ and $B^k$}
    Initialize parameters;

    \Repeat{Maximum iterations}{
      \textbf{Forward Propagation:}\;

     \Do{}{
      Conducting convolution operation with the raw vibration data using Eq. \eqref{eq:sigment_conv}.\;

      Use ReLU as the nonlinear activation function. \;

      Max-pooling function Eq. \eqref{eq:sigment_ply} is employed to extract the maximum feature values.\;

       }
       Conventional fully-connected layer is used for DEI regression.\;

       The sigmoid function is introduced for normalized output.\;

       Compute the MSE with the loss function using Eq. \eqref{eq:lossfx}. \;

       \textbf{Backward Propagation:}\;

       Compute the gradient using \textit{Adam} optimizer \cite{kingma2014adam} and update network parameters $W^k$ and $B^k$.
    }
    Use the trained CNN to estimate the DEI ${\bf{L_{test}}}$ on the test bearing.
    \caption{Outline of CNN training for DEI estimation}
\end{algorithm}


\subsection{RUL prediction}
\label{sec:SVR}
\begin{figure}[htb!]
              \centering                           
              \includegraphics[width=0.9\columnwidth]{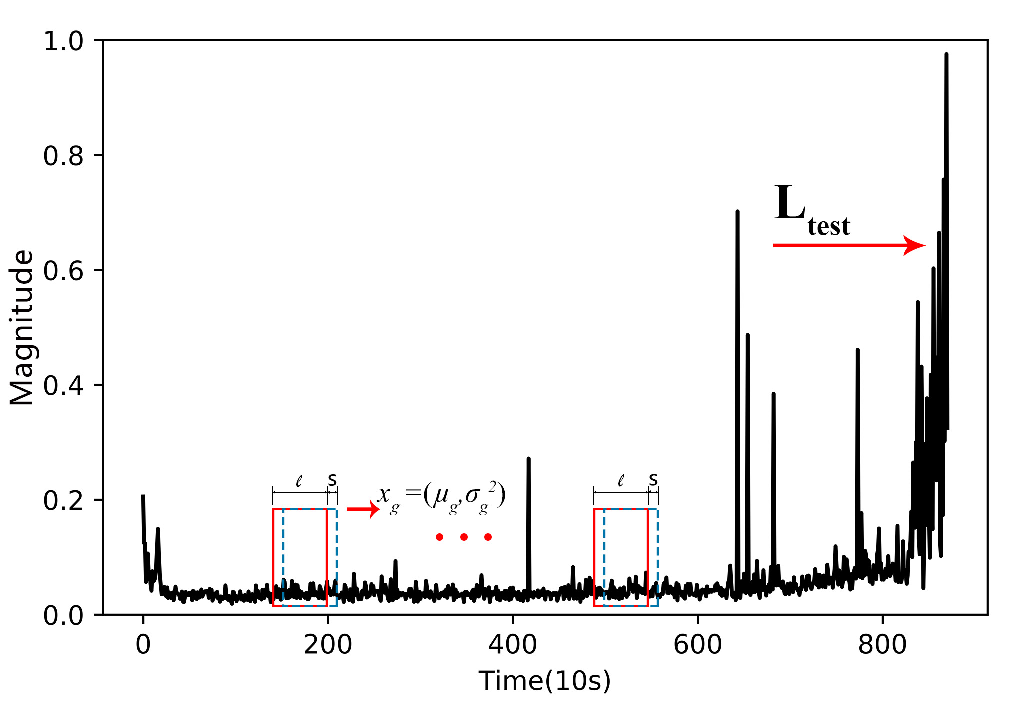}\vspace{-0.5em}
              \caption{The diagram of extracting features for $\epsilon$-SVR prediction, where $l$ is the sampling window size, $s$ is the sliding size, and $x_g$ stores the mean value $\mu_g$ and variance $\sigma_g^2$ of each sampling window. }
              \label{fig:SVR}
              \end{figure}

With the obtained CNN model, for a new test bearing with $Q\in \mathbb{Z}$ historical units, the estimated DEI ${\bf{L_{test}}}=[L_{1,test}, \ldots,L_{Q,test}] \in \mathbb{R}^{1\times Q}$ can be automatically generated by the trained CNN with the new vibration signal $\bf{S_{i}}$, where $i\in {D_{test}}=\left \{1,\ldots,Q\right \}$. Then, to predict the RUL $\widehat{T}_{failure}\in \mathbb{R}^+$, a $\epsilon$-SVR forecasting  model \cite{benkedjouh2013remaining} is formalized to predict the upcoming degradation $\widehat{L}_{Q+1,test}$, $\widehat{L}_{Q+2,test}$, $\ldots$ based on the estimated DEI $\bf{L_{test}}$ by sliding window method. The forecasting process contains three sub-steps:
\begin{enumerate}
\item Extract training features from the estimated DEI $\bf{L_{test}}$ over a sliding window. The schematic of this step is illustrated in Fig. \ref{fig:SVR}. The estimated DEI $\bf{L_{test}}$ is decomposed into overlapping windows associated with sampling window size $l$ and sliding size $s$. $x_g=(\mu_{g}, \sigma_{g}^2)$ for $g\in\{1,\ldots,\frac{Q-l-1}{s}+1\}$ represents a training feature for $\epsilon$-SVR, where $\mu$ denotes the mean value and $\sigma^2$ denotes the variance of each sampling window. Thus, the training set for $\epsilon$-SVR ${\bf{X}}=[(x_1,L_{l+1,test}),\ldots,(x_g,L_{(g-1)s+l+1,test}),\ldots,(x_{\frac{Q-l-1}{s}+1},$ $L_{Q,test})] $ is obtained, where $L_{(g-1)s+l+1,test}$ corresponds to the next value in $\bf{L_{test}}$ of the $g$-th sampling window;
\item $\epsilon$-SVR modeling is described in Algorithm \ref{alg:Framwork}, while at the application level, two parameters (distance limit $\epsilon\in \mathbb{R}$ and penalty parameter $C\in \mathbb{R}$) can be set manually when training the prediction model. A radial basis function (RBF) is necessary when we intend to train a nonlinear model;
\item The SVR model $f(x)$ learned in Algorithm \ref{alg:Framwork} is then used to predict the RUL by sliding window method (with the same $l$ and $s$ in step 1). Since the test bearing undergoing same operating condition as the training bearing, it is reasonable to define the FTs (denoted as $L_{ft}$) of the test bearing equals to the last feature of the DEI of the training bearing, such that $L_{ft}$=$L_N$. Hence, the first prediction can be calculated as $\widehat{L}_{Q+1,test}=f(x_{\frac{Q-l}{s}+1})$, and the predicted DEI ${\bf{\widehat{L}_{U,test}}}=[\widehat{L}_{Q+1,test},\ldots,\widehat{L}_{Q+U,test}]$ can be obtained by shifting the sampling window, with  $\widehat{L}_{Q+U-1,test}\leq L_{ft} \leq \widehat{L}_{Q+U,test}$. This will consequently lead to $\widehat{T}_{failure}= U\times \tau$, where $\tau$ is the time interval between two recording phases.
\end{enumerate}

\begin{algorithm}[]
\caption{ Framework of $\epsilon$-SVR.}
\label{alg:Framwork}
\begin{algorithmic}[1]
\REQUIRE ~~\\
A training set $\bf{X}$;\\
A distance limit $\epsilon$ and a penalty parameter $C$;\\ A kernel function named RBF with the equation: $\kappa(x_g,x_q) = \exp\left(-\frac{\left\|x_g-x_q\right\|^2}{2\sigma^2}\right)$, $\sigma$ is the width of RBF and $\sigma>0;$\\
\ENSURE ~~\\
 A regression model like $f({x})=w^T \phi({x})+b$, where $w$ and $b$ are optimized parameters and $\phi(x)$ is the $x$-mapped eigenvector that satisfies the equation: $\kappa(x_g,x_q)=\langle\phi(x_g),\phi(x_q)\rangle$.\\
 
\,\,\textbf{\textit{Step 1:}} Establish optimization problems using $C$, $\epsilon$, $\bf{X}$, and two slack variables $\xi_{g}$ and $\widehat{\xi}_{g}$ (slack degree for upper boundary and lower boundary, respectively) as following:
{\footnotesize \begin{alignat}{2}
  \min\limits_{w,b,\xi_{g},\widehat{\xi}_{g}}  \quad & \frac{1}{2}\left\|w\right\|^2+C\Sigma_{g=1}^{\frac{Q-l-1}{s}+1}(\xi_{g}+\widehat{\xi}_{g}) \tag{i}\label{eq:A1}\\
  \text{s.t.} \quad &  \begin{aligned}[t]
     f(\phi(x_g))-L_{(g-1)s+l+1,test} & \leq \epsilon+\xi_{g} \nonumber\\
     L_{(g-1)s+l+1,test}-f(\phi(x_g))  & \leq  \epsilon+\widehat{\xi}_{g}\\
     \xi_{g}\geq 0, \widehat{\xi}_{g}\geq 0.\,\,\,\,\,\,\,\,\,\,\,\,\,\,\,\,\,\,\,\,\,\,\,\,\,\,\,\,\,\,\,\,\,\,\,&   
  \end{aligned}
\end{alignat}}
\label{code:fram:trainbase}
\!\!\textbf{\textit{Step 2:}} Add Lagrangian multipliers for each constraint: $\mu_g\geq0,\widehat{\mu}_g\geq0,$ $\alpha_g \geq 0,\widehat{\alpha}_g\geq0$ and get the Lagrange function of formula  \eqref{eq:A1}:
{\footnotesize \begin{align}
 &L(w,b,\alpha,\widehat{\alpha},\xi,\widehat{\xi},\mu,\widehat{\mu})=\frac{1}{2}\left\|w\right\|^2 +C\Sigma_{g=1}^{\frac{Q-l-1}{s}+1}(\xi_{g}+\widehat{\xi}_{g})\nonumber\\
 &-\Sigma_{g=1}^{\frac{Q-l-1}{s}+1}\mu_g\xi_{g}-\Sigma_{g=1}^{\frac{Q-l-1}{s}+1}\widehat{\mu}_g\widehat{\xi}_{g}+\Sigma_{g=1}^{\frac{Q-l-1}{s}+1}\alpha_g(f(\phi(x_g))\nonumber \\
 &-L_{(g-1)s+l+1,test}-\epsilon-\xi_g) +\Sigma_{g=1}^{\frac{Q-l-1}{s}+1}\widehat{\alpha}_g(L_{(g-1)s+l+1,test}\nonumber\\
 &-f(\phi(x_g))-\epsilon-\widehat{\xi}_g) \,. \tag{ii}\label{eq:A2}
\end{align}}
\!\!\textbf{\textit{Step 3:}} Substitute $f(\bf{X})$ into the formula \eqref{eq:A2}, we firstly compute the gradient of $L(w,b,\alpha,\widehat{\alpha},\xi,\widehat{\xi},\mu,\widehat{\mu})$ with respect to $w, b, \xi_g$ and $\widehat{\xi}_g$ and make it equal to zero. Then get the following results: 
{\footnotesize \begin{align}
&w=\Sigma_{g=1}^{\frac{Q-l-1}{s}+1}(\widehat{\alpha}_g-\alpha_g)\phi(x_g), \quad 0=\Sigma_{g=1}^{\frac{Q-l-1}{s}+1}(\widehat{\alpha}_g-\alpha_g),\nonumber \\
&C=\alpha_g +\mu_g, \quad  \ C=\widehat{\alpha}_g +\widehat{\mu}_g. \tag{iii}\label{eq:A3}
 \end{align}}
\!\!\textbf{\textit{Step 4:}} Substitute formula \eqref{eq:A3} into formula \eqref{eq:A2}, and get a dual problem:
{\footnotesize \begin{alignat}{2}
\begin{aligned}  \max\limits_{\alpha,\widehat{\alpha}}\\\,  \end{aligned}\quad & \begin{aligned} &\Sigma_{i=g}^{\frac{Q-l-1}{s}+1}L_{(g-1)s+l+1,test}(\widehat{\alpha}_g-\alpha_g)-\epsilon(\widehat{\alpha}_g+\alpha_g) \nonumber \\
      &-\frac{1}{2}\Sigma_{g,q=1}^{\frac{Q-l-1}{s}+1}(\widehat{\alpha}_g-\alpha_g)(\widehat{\alpha}_{q}-\alpha_{q})\phi (x_g)^\top 
       \end{aligned}\\ 
  \text{s.t.} \quad &  \begin{aligned}[t]
     &\Sigma_{g=1}^{\frac{Q-l-1}{s}+1}(\widehat{\alpha}_g-\alpha_g)=0\nonumber\\
     &0\leq \alpha_g, \widehat{\alpha}_g\leq C.  
  \end{aligned}
\end{alignat}}
\!\!\textbf{\textit{Step 5:}} The above process satisfies the Karush–Kuhn–Tucker (KKT) condition, which means at least one of $\alpha_g$ and $\widehat{\alpha}_g$ is equal to $0$. A sequential minimization algorithm is used to solve $\alpha_g$ or $\widehat{\alpha}_g$. Parameter $b$ is calculated by substituting the $\alpha_g$ to get the mean value. The final regression model is described as:
{\footnotesize
\begin{align}
f(x)=\Sigma_{g=1}^{\frac{Q-l-1}{s}+1}(\widehat{\alpha}_g-\alpha_g)\kappa(x,x_g)+b.\nonumber
\end{align}}
\end{algorithmic}
\end{algorithm}

\section{Experiments}
\label{sec:tests}
\subsection{Data description}
\label{sec:datasets}
\begin{figure}[!htb]
\centering
\includegraphics[width=0.9\columnwidth]{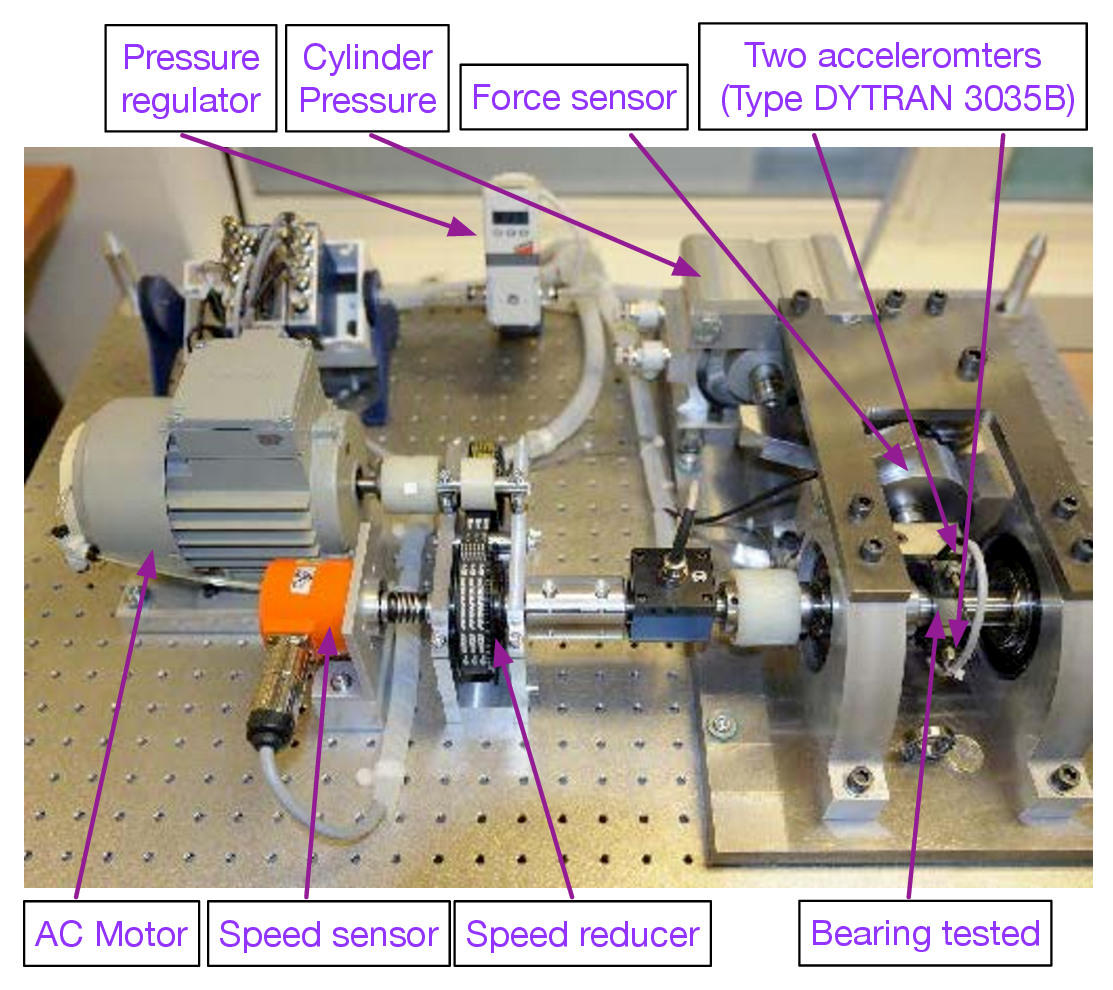}
\caption{Overview of the PRONOSTIA platform \cite{nectoux2012pronostia}.}
\label{fig:platform1}
\end{figure}

The validation of the proposed RUL prediction framework is conducted on an experimentation platform named PRONOSTIA (see Fig. \ref{fig:platform1}). This platform is built as a combination of three parts: a rotating part, a loading part, and a measurement part. The rotation of the test bearing is driven by the low speed shaft whose rotating torque is transmitted from an AC motor. A radial force generated by this loading part is applied on the external ring of the testing bearing. Since this external radial force exceeds the bearing's allowable dynamic load (4000N), the degradation behavior is accelerated so that we can observe its degradation process within a relatively short time in few hours. During experiment tests, two high-frequency accelerometers (Type DYTRAN 3035B) are placed orthogonally on the external race of the test bearing to acquire the horizontal and vertical vibrations respectively. The  accelerometer bandwidth is 0.5$\sim$10\,kHz ($\pm$ 5\%), and its natural/resonant frequency is 45\,kHz. In this work, we extract our degradation labels by using the horizontal vibrations.

Bearing1 is chosen to validate the proposed framework. More specifically, the training set bearing1\_2 is used for extracting the DEI and training the CNN model. Test sets bearing1\_4, bearing1\_5, and bearing1\_6 are then used for estimating their DEIs and predicting the RULs by applying $\epsilon$-SVR forecasting model. Results of the Bearing2 under different rotational frequency and external dynamic load are also provided and compared. The geometry parameters and the operation conditions of the bearing1 and bearing2 are listed in Table \ref{tb:parameters of bearing}. Sampling frequency of the vibration sensor is 25.6\,kHz. 0.1\,s accelerometer vibration signals are recorded at a fixed time interval $\tau= 10$\,s. Therefore, each recording phase contains $p=2560$ measurements. More detailed description of the data set, bearings, and sensors can refer to the data description in \cite{nectoux2012pronostia}.

\vspace{-1em}
 \begin{table}[hb!]
 \centering
 \caption{\label{tb:parameters of bearing}Physical characteristics and operation condition}
 \setlength{\tabcolsep}{8mm}
 \begin{tabular}{lc}
  \toprule
  Physical parameter & Value\\
  \hline
  \midrule
  Number of balls of the bearing ($\eta$)  & 13 \\
  Ball diameter of the bearing ($\ell_{ball}$)    & 3.5 mm\\
  Pitch diameter of the bearing ($\ell_{pitch}$)   & 25.6 mm \\
  Contact angle of the bearing ($\Phi$)  &  $0^\circ$  \\
  Rotation frequency ($f_\omega$), bearing1              & 1800 r/min\\
    Rotation frequency ($f_\omega$), bearing2              & 1600 r/min\\
  Maximum dynamic load (F), bearing1                & 4000 N  \\
  Maximum dynamic load (F), bearing2                & 4200 N  \\
  \bottomrule
 \end{tabular}
\end{table}

\vspace{0em}
\subsection{Degradation indicator extraction}
\label{sec:ft_extract}
\begin{figure}[htb!]
 \centering
\includegraphics[width=1\columnwidth]{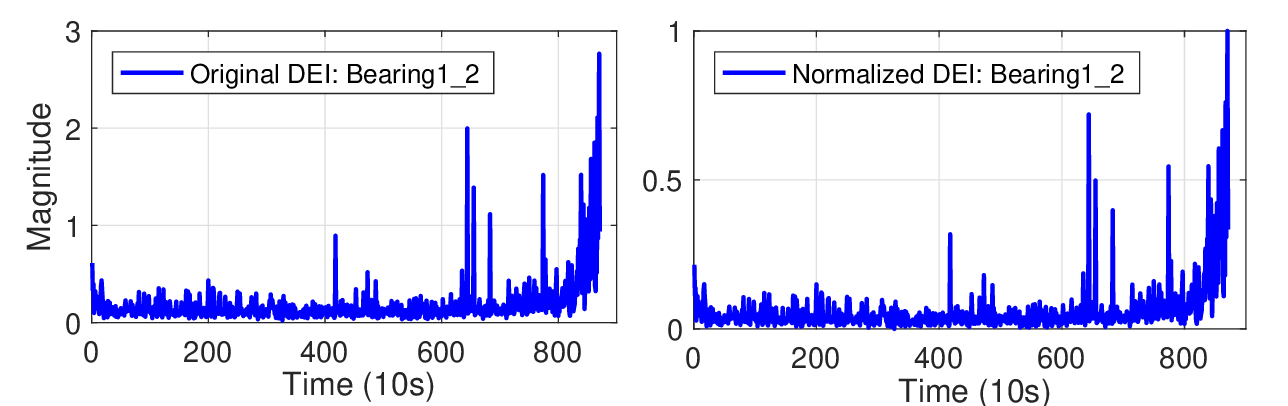}
\caption{Evolutions of (a) original DEI and (b) normalized DEI. }
\label{fig:rawDEI}
\end{figure}

\begin{figure*}[htb!]
\centering
\includegraphics[width=1.6\columnwidth]{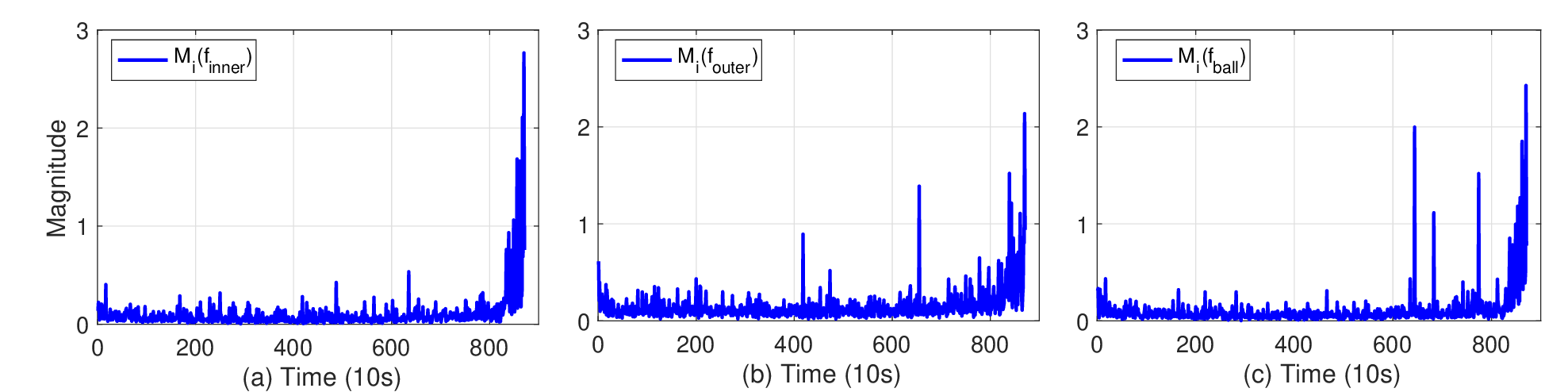}
\caption{ Time evolutions of the intermedia feature $M_i(f_{inner})$, $M_i(f_{outer})$, and $M_i(f_{ball})$ in Eq. \eqref{eq:mhs}.}
\label{fig:minner}
\end{figure*}

The DEI of the bearing1\_2 is extracted by substituting its outer ring frequency $f_{outer} $= 168\,Hz, inner ring frequency $f_{inner}$ = 221\,Hz, and ball frequency $f_{ball}$ = 215.4\,Hz into Eq. \eqref{eq:mhs}. The evolution of the final extracted DEI $\bf{L}$ of the bearing1\_2 is showed in Fig. \ref{fig:rawDEI}(a). We also show the time evolutions of intermediate features $M_i(f_{inner})$, $M_i(f_{outer})$, and $M_i(f_{ball})$ of Eq. \eqref{eq:mhs} in Fig. \ref{fig:minner}. It can be observed that the magnitude value of each time point in Fig. \ref{fig:rawDEI}(a) is the maximum of $M_i(f_{inner})$, $M_i(f_{outer})$, and $M_i(f_{ball})$ at that time. 

\subsection{Degradation indicator estimation}
\label{sec:ft_extract}

We use the vibration signal in the horizontal direction of the bearing1\_2 as the input of the CNN, and the normalized DEI is used as the label which contains historical units $N=871$.  The normalized DEI, $\bf{L_{norm}}$, is shown in Fig. \ref{fig:rawDEI}(b). A less complex architecture of the CNN model is designed to improve the robustness of the network. As shown in Fig. \ref{fig:cnnmodel}, our finalized CNN model consists of $K=6$ layers: two convolutional layers (Conv1 and Conv2), two max-pooling layers (Maxpooling1 and Maxpooling2), one fully connected layer (FC1), and one regression layer for output. Before model training, \textit{Adam} is set to be the optimizer with a small value 0.00001 as it guarantees a quick loss convergence compared with a larger or smaller learning rate for the CNN training process. The activation function in the output layer is the Sigmoid function, while ReLU function is used in the previous layers. Details of parameters in the proposed CNN model are concluded in Table \ref{tb:CNNpara}. The convolutional window sizes (kernel sizes) of convolutional layers are set to a large value 100 and a small value 2, respectively. The kernel size of Conv1 is relevantly large in order to extract more features from the raw vibration signal for more impressive power, meanwhile, small kernel size is selected for Conv2 to prevent overfitting. Hyper-parameters are obtained after 1000 iterations of training.

\begin{table}[b]
\centering
\caption{Parameters in the CNN model}
\label{tb:CNNpara}
\begin{tabular}{lccc}
\hline
Layer       & Filters & Kernel size/Stride & Output size \\ \hline
\hline
Input       & ...    & ...                & 1x2560x1    \\
Conv1       & 64     & 1x100/50           & 1x50x64     \\
Maxpooling1 & ...    & 1x2/2              & 1x25x64     \\
Conv2       & 64     & 1x2/1              & 1x24x64     \\
Maxpooling2 & ...    & 1x2/2              & 1x12x64     \\
Flatten     & ...    & ...                & 1x768       \\
FC1         & ...    & ...                & 1x100       \\
Output      & ...    & ...                & 1x1         \\ \hline
\end{tabular}
\end{table}

\begin{figure}[htb!]
              \centering
              \includegraphics[width=1\columnwidth]{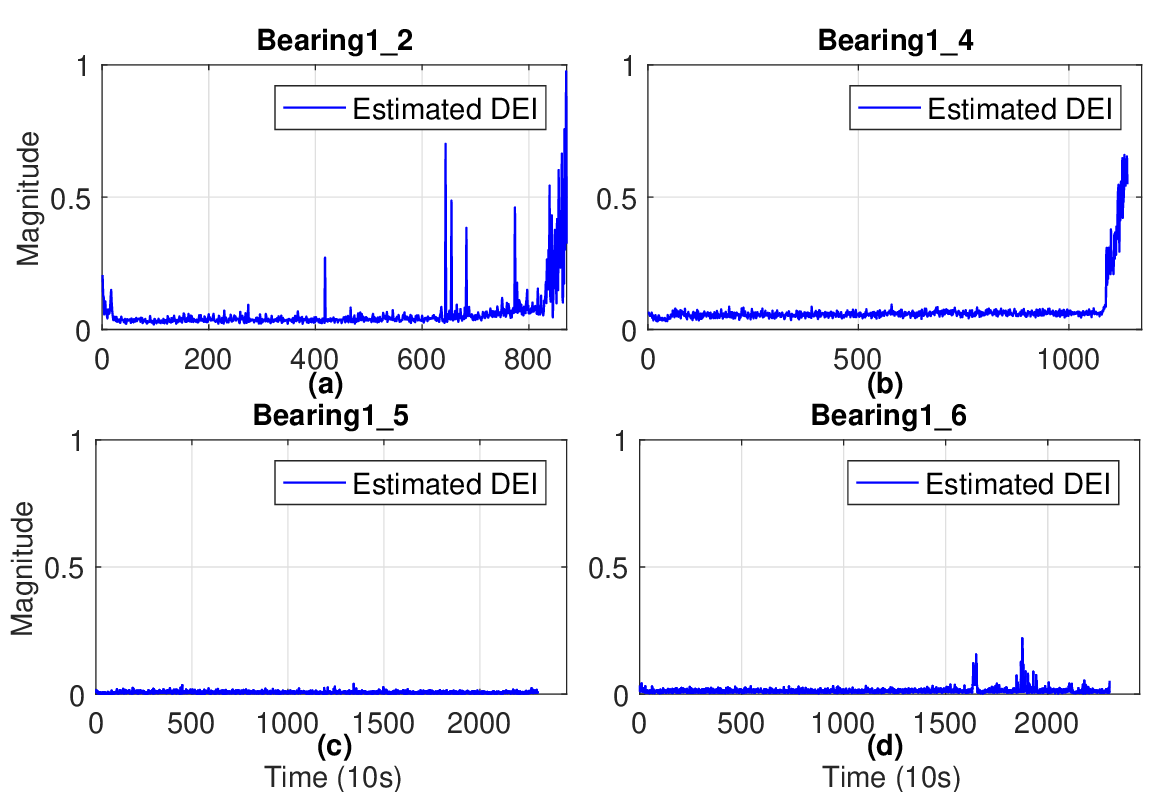}\\
              \caption{The estimated DEIs as the output of the proposed CNN model: (a) bearing1\_2, (b)bearing1\_4, (c) bearing1\_5, and (d) bearing1\_6. }
                \label{fig:online}
              \end{figure}

The estimated DEIs $\bf{L_{test}}$ as the output of the CNN model are shown in Fig. \ref{fig:online}. In Fig. \ref{fig:online}(a), estimated DEI of the training bearing1\_2 shows the similar time evolution as the DEI label in Fig. \ref{fig:rawDEI}(b). The final estimated DEI value of the bearing1\_2, $L_{ft}=0.9756$, is defined as the failure threshold for the test bearings in Fig. \ref{fig:online}(b)-(d).

\subsection{RUL prediction}
\label{sec:prediction}
As presented in Section \ref{sec:ft_extract}, the estimated DEIs have obtained from the trained CNN model. However, the DEIs of the test sets shown in  Fig. \ref{fig:online}(b)-(d) do not reach their fault limit, which need a regression algorithm to predict the estimated DEI till the end-of-life of each test bearing. A $\epsilon$-SVR method is proposed to predict the upcoming degradation process of the test bearings. We conduct a training on the predicted DEI of the bearing1\_2. The sampling window size $l$ and the moving size $s$ in Fig. \ref{fig:SVR} is set to 50 and 1, respectively. The kernel function used in the prediction case is an RBF and penalty parameter C of the error term is chosen as 5.09 by grid search and cross validation using the method of GridSearchCV \cite{pedregosa2011scikit}. We estimate the DEI after 1000 steps based on the existing DEI, and using the maximum value of predicted DEI of bearing1\_2 (i.e., $L_{tf}=0.9756$) to limit the termination time of the test sets.

\begin{figure}[htb!]
              \centering
              \includegraphics[width=1\columnwidth]{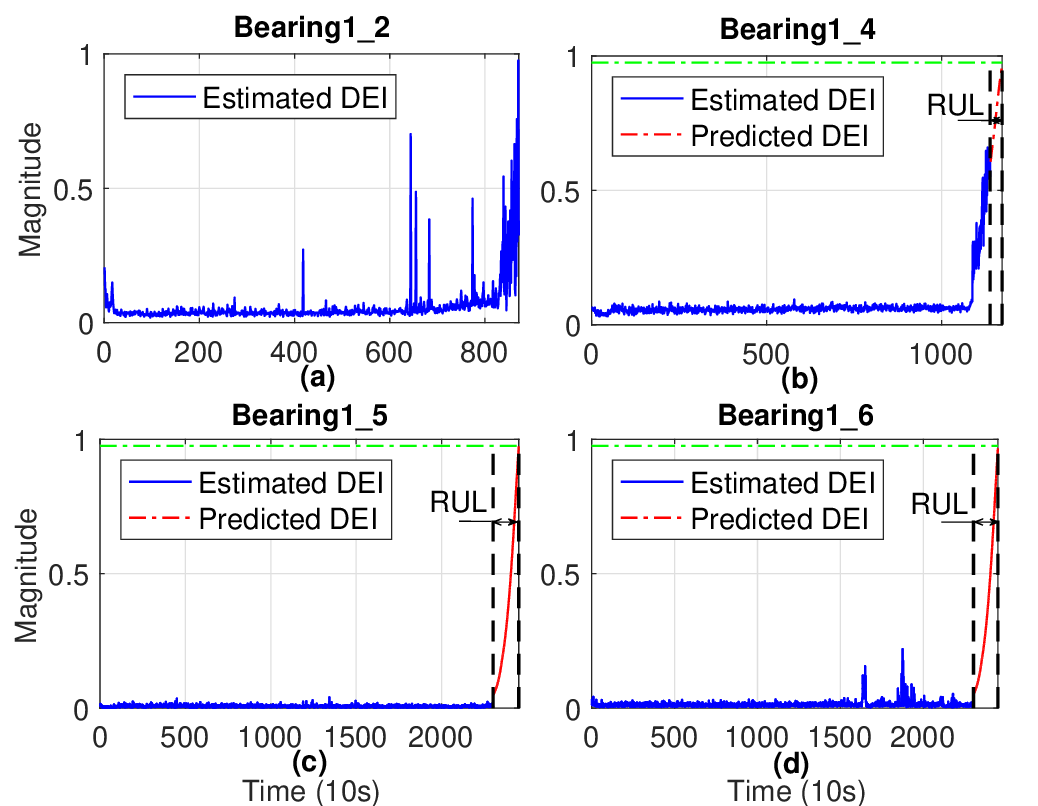}\\
              \caption{Plot (a) shows the estimated DEI of the bearing1\_2. Predicted RULs of the test bearings are (b) 340\,s for bearing1\_4, (c) 1500\,s for bearing1\_5, and (d) 1480\,s for bearing1\_6. Green dotted lines in (b)-(d) represent the FTs. }
                \label{fig:bearingmerge}
              \end{figure}

Fig. \ref{fig:bearingmerge}(b)-(d) show the predicted RULs $\widehat{T}_{failure}$ of the test bearings till the failures occur. Red lines represent the predicted evolution of the bearings' degradation behavior using the $\epsilon$-SVR method. The RULs are calculated as the difference between the final time when DEI reaching the failure threshold and the time of the last known point of the test bearings. For bearing1\_4, the predicted RUL is 340s, while 1500s and 1480s are the predicted RULs for bearing1\_5 and bearing1\_6, respectively.

\subsection{Comparison and discussion}
\label{sec:comparison}
To assess the accuracy of the proposed method and compare to other existing approaches, two metrics are commonly adopted: 1) The relative percentage error ($Er\%$) which is given by Eq. \eqref{eq:RUL}; and 2) The exponential transformed accuracy (ETA) proposed in IEEE PHM 2012 \cite{nectoux2012pronostia}. ETA is an assessment index to distinguish the seriousness of the underestimate and overestimate of RUL prediction. It is clearly that underestimate (i.e., early warning) is preferred than overestimation (i.e., warning after damage) to prevent more severe damage of the bearing. The formulas are expressed in Eq. \eqref{eq:ETA}.
                \begin{equation}
                       E_r=100\% \times \frac{T_{failure}-\widehat{T}_{failure}}{T_{failure}}
                       \label{eq:RUL}
                \end{equation}
                \vspace{-0.5em}
                 \begin{equation}
                         \textup{ETA}=\left\{\begin{matrix}
                         \exp(-\ln(0.5)\frac{E_r}{5})\,\,\,\,\,\,\,if \,\,E_r\leq 0 \\
                         \exp(+\ln(0.5)\frac{E_r}{20})\,\,\,\,\,\,\,if \,\,E_r> 0
                          \end{matrix}\right.
                \label{eq:ETA}
                \end{equation}
where where $T_{failure}$ is the real RUL for the test bearing. A higher $\mid$$Er\%$$\mid$ means a worse RUL prediction result. On the other hand, ETA value varies from 0 to 1, and a higher score means a better RUL prediction result. In this work, $Er$ is the common choice of most previous literature, thus it will be used for comparison.

In addition to the two metrics that evaluate the prediction performance for a specific bearing, three more assessment metrics, namely average score $S_{mean}$, mean average error MAE, and the normalized root mean square error NRMSE, to make a comprehensive comparison of different methods, which are with the expression forms
\begin{equation}
  S_{mean}=\frac{1}{\beta} \sum^{\beta}_{i=1} \textup{ETA}_i
\end{equation}
\vspace{-1em}
\begin{equation}
  \textup{MAE}=\frac{1}{\beta} \sum^{\beta}_{i=1} \mid T_{failure_i}-\hat{T}_{failure_i}\mid
\end{equation} 
\begin{equation}
  \textup{NRMSE}=\frac{\sqrt{\frac{1}{\beta}\sum^{\beta}_{i=1} (T_{failure_i}-\hat{T}_{failure_i})^2}}{\frac{1}{\beta}\sum^{\beta}_{i=1}\hat{T}_{failure_i}} 
\end{equation}
where $\beta$ is the number of test bearings.

To verify the benefits of the DEI and CNN techniques on RUL prediction, here we also develop other two tested methods for comparison purpose (see Table \ref{tb:method_cp}). The proposed method and the tested methods are denoted and explained as follows:
\begin{enumerate}
\item \textit{C1: CNN and $\epsilon$-SVR}: This tested method uses a conventional linear time degradation label for CNN training rather than the nonlinear DEI. By this means, we can illustrate the impact of the DEI on RUL prediction.
\item \textit{C2: DEI and $\epsilon$-SVR}: Without training the CNN model for feature extraction, DEI in this tested method is extracted manually and the $\epsilon$-SVR is followed for the RUL prediction. Note that computing a DEI for a new bearing requires high computational power and longer time. In the meantime, the FT of each test bearing has to be pre-defined artificially, which increase uncertainties of the RUL prediction affected by different working conditions. By this means, we can illustrate the impact of CNN modeling on the prediction of the final RUL.
\item \textit{Proposed method: DEI-based CNN and $\epsilon$-SVR}: This is the proposed framework which integrates DEI extraction, CNN, and $\epsilon$-SVR into one framework.
\end{enumerate}
\begin{table}[h]
\centering
\caption{Proposed method and tested methods for comparison}
\label{tb:method_cp}
\begin{tabular}{lcccc}
  \toprule
                                        & DEI & CNN                       & SVR                 \\ \hline
  \midrule
\textit{\textbf{C1}}           & -  & \checkmark & \checkmark \\
\textit{\textbf{C2}}           & \checkmark   & -                         & \checkmark   \\
\textit{\textbf{Proposed method}} & \checkmark   & \checkmark & \checkmark  \\
  \bottomrule
\end{tabular}
\end{table}

\begin{figure}[htb!]
              \centering
              \includegraphics[width=1\columnwidth]{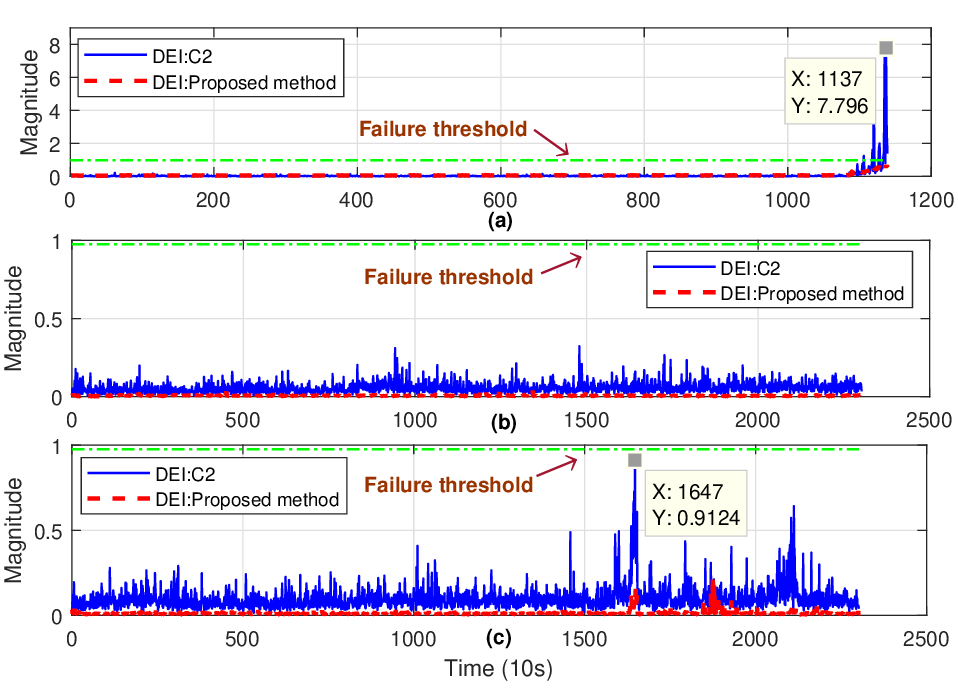}\\
              \caption{The comparison of the estimated DEIs extracted using HHT (DEI:C2) and calculated by trained CNN model (DEI:Proposed method) of (a) test bearing1\_4, (b) test bearing1\_5, and (c) test bearing1\_6.}
                \label{fig:DEIcomp}
              \end{figure}

The predicted numerical errors of the test bearings with the proposed approach and the tested methods are listed in Table \ref{comparison1}.  Our approach achieves $Er\%$ of -0.29\%, 6.83\%, and -1.37\% for bearing1\_4, bearing1\_5, and bearing1\_6, respectively, which are much more smaller than the C1 and C2 methods. C1 uses a linear time degradation label for the training of the CNN model. The results show more than 19\% prediction errors for test bearings and even 91.15\% prediction error is obtained from bearing1\_4, indicating that time degradation label is less effective than the DEI for the CNN training process. C2 is the method extracting the degradation indicator of the test bearings and define the FTs manually. As testing bearing1\_4, 1\_5, and 1\_6 operate under same working conditions of bearing1\_2, we employ the maximum and minimum value of bearing1\_2 to normalize the extracted DEIs of C2 method. With same working condition and same normalization parameters, FT of the proposed method could be reasonablely used in C2 as well. To evaluate the impact of CNN modeling on the estimation of the final RUL, for the test bearings, the estimated DEIs extracted using HHT and calculated by trained CNN model are compared in Fig. \ref{fig:DEIcomp}. Without the CNN modeling procedure, one of the main drawback of the C2 method is that it requires long time to calculate ($\sim$2\,s of each sampling period). This limits the practical application of this method in industry. Moreover, Fig. \ref{fig:DEIcomp}(a) shows that due to the uncertainties and huge mount of noise, the DEI extracted by C2 method has already exceeded the FT before the exact failure time, resulting in a 100\% $E_r$. Similarly, in Fig. \ref{fig:DEIcomp}(c), the DEI extracted of C2 method is much noised than it of proposed method. At 16470\,s, the magnitude of DEI:C2 is almost close to the FT, this phenomenon might lead to a waste of sources due to much underestimated of RUL. The comparison results in Table \ref{comparison1} demonstrate the benefits of using DEI and CNN in estimating RULs of REBs.

                \begin{table*}[]
                 \centering
                 \caption{\label{comparison1}Comparison of predicted results between proposed approach and other tested methods}\vspace{-0.5em}
                 \begin{tabular}{lccccccccccc}
                  \toprule

                        &   \multicolumn{3}{c}{Bearing1\_4  ($T_{failure}$ =339\,s)} &  & \multicolumn{3}{c}{Bearing1\_5 ($T_{failure}$ =1610\,s)} & & \multicolumn{3}{c}{Bearing1\_6 ($T_{failure}$ =1460\,s)}\\ \cline{2-4} \cline{6-8} \cline{10-12}
                     Methods     & $\widehat{T}_{failure}$ &$E_r\%$ &ETA &  & $\widehat{T}_{failure}$ &$E_r\%$&ETA & & $\widehat{T}_{failure}$ & $E_r\%$ &ETA\\
                  \hline
                  \midrule
                  \textbf{Proposed approach}             & 340\,s   &\textbf{-0.29}\% &\textbf{0.96}  &   & 1500\,s     & \textbf{6.83}\% &\textbf{0.79}   &   & 1480\,s     & \textbf{-1.37}\%  &\textbf{0.83}   \\
                  C1      & 30\,s    &91.15\%   &0.04   &   & 820\,s    & 49.07\%  &0.18   &   & 1181\,s   & 19.11\% &0.52     \\
                  C2        & N/A (0\,s)   & N/A (100\%)   &0.03   &   & 1140\,s   & 29.19\%  &0.36   &   & 1080\,s   & 26.02\%  &0.41   \\
                  \bottomrule
                 \end{tabular}
                \end{table*}

                 \begin{table*}[htb!]
                 \centering
                 \caption{Comparison of predicted results between proposed approach and existing major approaches.}\vspace{-0.5em}
                 
\begin{tabular}{lcccccccc}
\hline
                                            & \multicolumn{5}{c}{$E_r\%$}                                                                                                                                                    & \multicolumn{1}{c}{\multirow{2}{*}{$S_{mean}$}} & \multirow{2}{*}{MAE} & \multirow{2}{*}{NRMSE} \\ \cline{2-6}
Methods                                     & Bearing1\_4                       & Bearing1\_5                      & Bearing1\_6                       & Bearing2\_4                      & Bearing 2\_6                     & \multicolumn{1}{c}{}                            &                      &                        \\ \hline
\textbf{Proposed approach} & \textbf{-0.29}\% & \textbf{6.83}\% & \textbf{-1.37}\% & \textbf{5.75}\% & \textbf{1.55}\% &    \textbf{0.87}                                             &    \textbf{46.2}                  &  \textbf{0.05}                      \\
FCAMN \cite{chen2019real}                                       & 21.95\%                           & -15.22\%                         & -5.74\%                           & -                                & -                                &   0.35                                              &      696.7               &        0.13                \\
Multi-scale CNN   \cite{zhu2018estimation}                          & 10.69\%                           & -148.20\%                        & -21.51\%                          & -                                & -                                &  0.25                                               &        1003.3              &     0.50                   \\
CWT-CNN      \cite{yoo2018novel}                                & 20.35\%                           & 11.18\%                          & 34.93\%                           & -1.44\%                          & -42.64\%                         & 0.46                                                &        265.8              &     0.29                   \\
RNN  \cite{guo2017recurrent}                                        & 62.07\%                           & -22.98\%                         & 21.23\%                           & -19.42\%                         & -13.95\%                         &  0.17                                               &        586.0             &        0.57                \\
Particle Filtering     \cite{lei2016model}                     & 5.60\%                            & 100.00\%                         & 28.08\%                           & 8.63\%                           & 58.91\%                          &  0.42                                               &        583.8              &        1.29                \\
FFT+Ratio      \cite{sutrisno2012estimation}      & 80.00\%                           & 9.00\%                           & -5.00\%                           & 10\%                             & 49\%                             &  0.44                                               &         252             &        0.32                \\
LSTM    \cite{hinchi2018rolling}      & 38.69\%                           & -99.40\%                         & -120.07\%                         & 19.81\%                          & 17.87\%                          &   0.26                                              &   996.2                &     0.57                   \\
SOM   \cite{hong2014condition}     & -20.94\%                          & -278.26\%                        & 19.18\%                           & 51.80\%                          & -20.93\%                         &  0.16                                              &   1164.2                   &      1.02                  \\ \hline
\end{tabular}
\label{tb:comparison}
\end{table*}

 To demonstrate the generality of our proposed RUL estimation framework, other test bearings (i.e, bearing2\_4 and bearing2\_6) that operates undergoing a different external load and rotational speed are analyzed. Bearing2\_2 is used for training, and the trained CNN model is obtained without changing any hyper-parameters and the architecture of model in Table \ref{tb:CNNpara}. We just fine-tune the $\epsilon$-SVR forecasting model by changing penalty parameter C from 5.09 to 7.09, resulting in 5.75\% and 1.55\% $E_r$ for Bearing2\_4 and bearing2\_6, respectively. Good RUL prediction results of bearings with different operating conditions indicate the repeatability and robustness of our proposed method, with respect to the hyper parameters and the architecture of the CNN.

To further verify the proposed approach, the predicted numerical errors of RULs generated by the proposed method and eight published methods are compared and listed in Table \ref{tb:comparison}.  Other published approaches include a recurrent neural network method based health indicator \cite{guo2017recurrent}, the method proposed by the winner of the IEEE PHM 2012 prognostic \cite{sutrisno2012estimation}, a convolutional long-short-term memory network (LSTM) method \cite{hinchi2018rolling}, and a self-organization Mapping (SOM) method, etc.. In addition, recent CNN-based approaches, including the frozen convolution and activated memory network (FCAMN) \cite{chen2019real}, multi-scale CNN \cite{zhu2018estimation}, and continuous wavelet transform CNN (CWT-CNN)  \cite{yoo2018novel} are also compared. The results of the comparison shown in Table \ref{tb:comparison} confirm that our approach significantly outperforms the referenced methods with the highest average score $S_{mean}=0.87$, and the smallest MAE\,=\,46.2 and NRMSE\,=\,0.05. In particular, a -0.29\% $E_{r}$ of bearing1\_4 is achieved, owing to a 1s absolute time error. This result benefits from a good nonlinear degradation indicator extracted using the HHT method.  In addition, the CNN is a powerful tool for discovering the hidden pattern of the extracted degradation indicator and the underlying bearing system, further increasing the accuracy of the predicted RUL.

It can be concluded from the experiment results that the proposed data-driven RUL estimation approach has much smaller prediction errors, compared with both the tested methods in this work and the other published methods in previous studies.

\vspace{-0.5em}
\section{Conclusion}
\label{sec:conclusion}
In this paper, a data-driven framework for RUL prediction of rolling element bearing is presented using the HHT method, a CNN model, and an $\epsilon$-SVR forecasting model. A nonlinear degradation indicator DEI is first extracted from the raw vibration signals using the HHT method, which is defined as the label for the training. A CNN model is trained to discover the hidden pattern between the extracted DEI and the raw vibration data of the training bearing. In this way, predicted DEIs are automatically obtained when applying the trained CNN model to the test bearings. Finally, the RULs of the testing bearings are obtained using an $\epsilon$-SVR forecasting model. An experimentation platform that allows to observe the accelerated degradation process of bearings is employed to validate the proposed framework. The proposed framework achieves much smaller prediction errors for RUL predictions than previous published approaches.

Future work includes the application of the proposed framework to a wider range of case studies on experimental data in other applications \cite{yuan2019data}, and the investigation of other potential degradation labels to achieve even higher accuracy in estimating RUL.

\bibliographystyle{IEEEtran}

\begin{thebibliography}{10}
\providecommand{\url}[1]{#1}
\csname url@samestyle\endcsname
\providecommand{\newblock}{\relax}
\providecommand{\bibinfo}[2]{#2}
\providecommand{\BIBentrySTDinterwordspacing}{\spaceskip=0pt\relax}
\providecommand{\BIBentryALTinterwordstretchfactor}{4}
\providecommand{\BIBentryALTinterwordspacing}{\spaceskip=\fontdimen2\font plus
\BIBentryALTinterwordstretchfactor\fontdimen3\font minus
  \fontdimen4\font\relax}
\providecommand{\BIBforeignlanguage}[2]{{%
\expandafter\ifx\csname l@#1\endcsname\relax
\typeout{** WARNING: IEEEtran.bst: No hyphenation pattern has been}%
\typeout{** loaded for the language `#1'. Using the pattern for}%
\typeout{** the default language instead.}%
\else
\language=\csname l@#1\endcsname
\fi
#2}}
\providecommand{\BIBdecl}{\relax}
\BIBdecl

\bibitem{harris2001rolling}
T.~A. Harris, \emph{Rolling bearing analysis}.\hskip 1em plus 0.5em minus
  0.4em\relax John Wiley and sons, 2001.

\bibitem{yuan2017bayesian}
Y.~Yuan, H.-T. Zhang, Y.~Wu, T.~Zhu, and H.~Ding, ``Bayesian learning-based
  model-predictive vibration control for thin-walled workpiece machining
  processes,'' \emph{IEEE/ASME Transactions on Mechatronics}, vol.~22, no.~1,
  pp. 509--520, 2017.

\bibitem{heng2009rotating}
A.~Heng, S.~Zhang, A.~C. Tan, and J.~Mathew, ``Rotating machinery prognostics:
  State of the art, challenges and opportunities,'' \emph{Mechanical Systems
  and Signal Processing}, vol.~23, no.~3, pp. 724--739, 2009.

\bibitem{rodrigues2018remaining}
L.~R. Rodrigues, ``Remaining useful life prediction for multiple-component
  systems based on a system-level performance indicator,'' \emph{IEEE/ASME
  Transactions on Mechatronics}, vol.~23, no.~1, pp. 141--150, 2018.

\bibitem{de2016data}
I.~V. de~Bessa, R.~M. Palhares, M.~F. S.~V. D'Angelo, and J.~E. Chaves~Filho,
  ``Data-driven fault detection and isolation scheme for a wind turbine
  benchmark,'' \emph{Renewable Energy}, vol.~87, pp. 634--645, 2016.

\bibitem{wang2018real}
Z.-Q. Wang, C.-H. Hu, and H.-D. Fan, ``Real-time remaining useful life
  prediction for a nonlinear degrading system in service: Application to
  bearing data,'' \emph{IEEE/ASME Transactions on Mechatronics}, vol.~23,
  no.~1, pp. 211--222, 2018.

\bibitem{li2000stochastic}
Y.~Li, T.~Kurfess, and S.~Liang, ``Stochastic prognostics for rolling element
  bearings,'' \emph{Mechanical Systems and Signal Processing}, vol.~14, no.~5,
  pp. 747--762, 2000.

\bibitem{si2011remaining}
X.-S. Si, W.~Wang, C.-H. Hu, and D.-H. Zhou, ``Remaining useful life
  estimation--a review on the statistical data driven approaches,''
  \emph{European Journal of Operational Research}, vol. 213, no.~1, pp. 1--14,
  2011.

\bibitem{wang2019multi}
J.~Wang, P.~Fu, L.~Zhang, R.~X. Gao, and R.~Zhao, ``Multi-level information
  fusion for induction motor fault diagnosis,'' \emph{IEEE/ASME Transactions on
  Mechatronics}, 2019.

\bibitem{gao2015feature}
H.~Gao, L.~Liang, X.~Chen, and G.~Xu, ``Feature extraction and recognition for
  rolling element bearing fault utilizing short-time fourier transform and
  non-negative matrix factorization,'' \emph{Chinese Journal of Mechanical
  Engineering}, vol.~28, no.~1, pp. 96--105, 2015.

\bibitem{yan2014wavelets}
R.~Yan, R.~X. Gao, and X.~Chen, ``Wavelets for fault diagnosis of rotary
  machines: A review with applications,'' \emph{Signal Processing}, vol.~96,
  pp. 1--15, 2014.

\bibitem{meng1991rotating}
Q.~Meng and L.~Qu, ``Rotating machinery fault diagnosis using wigner
  distribution,'' \emph{Mechanical Systems and Signal Processing}, vol.~5,
  no.~3, pp. 155--166, 1991.

\bibitem{soualhi2015bearing}
A.~Soualhi, K.~Medjaher, and N.~Zerhouni, ``Bearing health monitoring based on
  hilbert--huang transform, support vector machine, and regression,''
  \emph{IEEE Transactions on Instrumentation and Measurement}, vol.~64, no.~1,
  pp. 52--62, 2015.

\bibitem{wu2018degradation}
J.~Wu, C.~Wu, S.~Cao, S.~W. Or, C.~Deng, and X.~Shao, ``Degradation data-driven
  time-to-failure prognostics approach for rolling element bearings in
  electrical machines,'' \emph{IEEE Transactions on Industrial Electronics},
 vol.~66, no.~1, pp. 529--539, 2018.

\bibitem{cheng2013gear}
G.~Cheng, Y.-l. Cheng, L.-h. Shen, J.-b. Qiu, and S.~Zhang, ``Gear fault
  identification based on hilbert--huang transform and som neural network,''
  \emph{Measurement}, vol.~46, no.~3, pp. 1137--1146, 2013.

\bibitem{tian2012artificial}
Z.~Tian, ``An artificial neural network method for remaining useful life
  prediction of equipment subject to condition monitoring,'' \emph{Journal of
  Intelligent Manufacturing}, vol.~23, no.~2, pp. 227--237, 2012.

\bibitem{wang2007adaptive}
W.~Wang, ``An adaptive predictor for dynamic system forecasting,''
  \emph{Mechanical Systems and Signal Processing}, vol.~21, no.~2, pp.
  809--823, 2007.

\bibitem{sikorska2011prognostic}
J.~Sikorska, M.~Hodkiewicz, and L.~Ma, ``Prognostic modelling options for
  remaining useful life estimation by industry,'' \emph{Mechanical Systems and
  Signal Processing}, vol.~25, no.~5, pp. 1803--1836, 2011.

\bibitem{hinton2012deep}
G.~Hinton, L.~Deng, D.~Yu, G.~E. Dahl, A.-r. Mohamed, N.~Jaitly, A.~Senior,
  V.~Vanhoucke, P.~Nguyen, T.~N. Sainath \emph{et~al.}, ``Deep neural networks
  for acoustic modeling in speech recognition: The shared views of four
  research groups,'' \emph{IEEE Signal Processing Magazine}, vol.~29, no.~6,
  pp. 82--97, 2012.

\bibitem{sun2019deep}
M.~{Sun}, I.~{Konstantelos}, and G.~{Strbac}, ``A deep learning-based feature
  extraction framework for system security assessment,'' \emph{IEEE
  Transactions on Smart Grid}, vol.~10, no.~5, pp. 5007--5020, Sep. 2019.

\bibitem{lecun2015deep}
Y.~LeCun, Y.~Bengio, and G.~Hinton, ``Deep learning,'' \emph{Nature}, vol. 521,
  no. 7553, p. 436, 2015.

\bibitem{yuan2019nsr}
Y.~Yuan, G.~Ma, C.~Cheng, B.~Zhou, H.~Zhao, H.-T. Zhang, and H.~Ding, ``A
  general end-to-end diagnosis framework for manufacturing systems,''
  \emph{National Science Review}, to appear, 2019.

\bibitem{chen2019real}
Z.~Chen, X.~Tu, Y.~Hu, and F.~Li, ``Real-time bearing remaining useful life
  estimation based on the frozen convolutional and activated memory neural
  network,'' \emph{IEEE Access}, vol.~7, pp. 96583--96593, 2019.

\bibitem{zhu2018estimation}
J.~Zhu, N.~Chen, and W.~Peng, ``Estimation of bearing remaining useful life
  based on multiscale convolutional neural network,'' \emph{IEEE Transactions
  on Industrial Electronics}, vol.~66, no.~4, pp. 3208--3216, 2018.

\bibitem{ren2018prediction}
L.~Ren, Y.~Sun, H.~Wang, and L.~Zhang, ``Prediction of bearing remaining useful
  life with deep convolution neural network,'' \emph{IEEE Access}, vol.~6, pp.
  13\,041--13\,049, 2018.

\bibitem{sun2020using}
M.~{Sun}, T.~{Zhang}, Y.~{Wang}, G.~{Strbac}, and C.~{Kang}, ``Using bayesian
  deep learning to capture uncertainty for residential net load forecasting,''
  \emph{IEEE Transactions on Power Systems}, vol.~35, no.~1, pp. 188--201, Jan
  2020.

\bibitem{lei2018machinery}
Y.~Lei, N.~Li, L.~Guo, N.~Li, T.~Yan, and J.~Lin, ``Machinery health
  prognostics: A systematic review from data acquisition to rul prediction,''
  \emph{Mechanical Systems and Signal Processing}, vol. 104, pp. 799--834,
  2018.

\bibitem{tandon1999review}
N.~Tandon and A.~Choudhury, ``A review of vibration and acoustic measurement
  methods for the detection of defects in rolling element bearings,''
  \emph{Tribology International}, vol.~32, no.~8, pp. 469--480, 1999.

\bibitem{neyshabur2017exploring}
B.~Neyshabur, S.~Bhojanapalli, D.~McAllester, and N.~Srebro, ``Exploring
  generalization in deep learning,'' in \emph{Advances in Neural Information
  Processing Systems}, 2017, pp. 5947--5956.

\bibitem{kingma2014adam}
D.~P. Kingma and J.~Ba, ``Adam: a method for stochastic optimization. corr
  abs/1412.6980 (2014),'' 2014.

\bibitem{benkedjouh2013remaining}
T.~Benkedjouh, K.~Medjaher, N.~Zerhouni, and S.~Rechak, ``Remaining useful life
  estimation based on nonlinear feature reduction and support vector
  regression,'' \emph{Engineering Applications of Artificial Intelligence},
  vol.~26, no.~7, pp. 1751--1760, 2013.

\bibitem{nectoux2012pronostia}
P.~Nectoux, R.~Gouriveau, K.~Medjaher, E.~Ramasso, B.~Chebel-Morello,
  N.~Zerhouni, and C.~Varnier, ``Pronostia: An experimental platform for
  bearings accelerated degradation tests.'' in \emph{IEEE International
  Conference on Prognostics and Health Management, PHM'12.}\hskip 1em plus
  0.5em minus 0.4em\relax IEEE Catalog Number: CPF12PHM-CDR, 2012, pp. 1--8.

\bibitem{pedregosa2011scikit}
F.~Pedregosa, G.~Varoquaux, A.~Gramfort, V.~Michel, B.~Thirion, O.~Grisel,
  M.~Blondel, P.~Prettenhofer, R.~Weiss, V.~Dubourg \emph{et~al.},
  ``Scikit-learn: Machine learning in python,'' \emph{{Journal of Machine
  Learning Research}}, vol.~12, no. Oct, pp. 2825--2830, 2011.

\bibitem{yoo2018novel}
Y.~Yoo and J.-G. Baek, ``A novel image feature for the remaining useful
  lifetime prediction of bearings based on continuous wavelet transform and
  convolutional neural network,'' \emph{Applied Sciences}, vol.~8, no.~7, p.
  1102, 2018.

\bibitem{guo2017recurrent}
L.~Guo, N.~Li, F.~Jia, Y.~Lei, and J.~Lin, ``A recurrent neural network based
  health indicator for remaining useful life prediction of bearings,''
  \emph{Neurocomputing}, vol. 240, pp. 98--109, 2017.

\bibitem{lei2016model}
Y.~Lei, N.~Li, S.~Gontarz, J.~Lin, S.~Radkowski, and J.~Dybala, ``A model-based
  method for remaining useful life prediction of machinery,'' \emph{IEEE
  Transactions on Reliability}, vol.~65, no.~3, pp. 1314--1326, 2016.

\bibitem{sutrisno2012estimation}
E.~Sutrisno, H.~Oh, A.~S.~S. Vasan, and M.~Pecht, ``Estimation of remaining
  useful life of ball bearings using data driven methodologies,'' in
  \emph{Prognostics and Health Management (PHM), 2012 IEEE Conference
  on}.\hskip 1em plus 0.5em minus 0.4em\relax IEEE, 2012, pp. 1--7.

\bibitem{hinchi2018rolling}
A.~Z. Hinchi and M.~Tkiouat, ``Rolling element bearing remaining useful life
  estimation based on a convolutional long-short-term memory network,''
  \emph{Procedia Computer Science}, vol. 127, pp. 123--132, 2018.

\bibitem{hong2014condition}
S.~Hong, Z.~Zhou, E.~Zio, and K.~Hong, ``Condition assessment for the
  performance degradation of bearing based on a combinatorial feature
  extraction method,'' \emph{Digital Signal Processing}, vol.~27, pp. 159--166,
  2014.

\bibitem{yuan2019data}
Y.~Yuan, X.~Tang, W.~Zhou, W.~Pan, X.~Li, H.-T. Zhang, H.~Ding, and
  J.~Goncalves, ``Data driven discovery of cyber physical systems,''
  \emph{Nature Communications}, vol.~10, no.~1, pp. 1--9, 2019.
  
\end{thebibliography}

\begin{IEEEbiography}[{\includegraphics[width=1in,height=1.25in,clip,keepaspectratio]{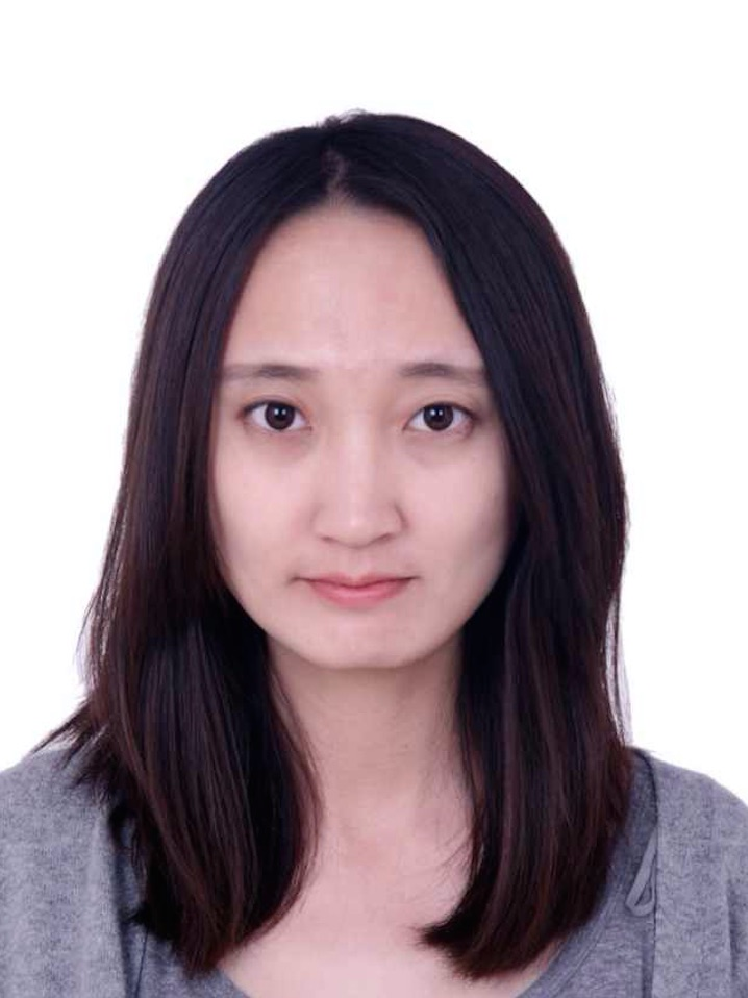}}]{Cheng Cheng} received the B. Eng. in Measurement, Control Technology and Instrument in 2012 from Tianjin University, China. In 2013 and 2018, she respectively received the MSc and the Ph.D. in Control Systems from Imperial College London, UK. Since 2018, she has been a Postdoctoral Researcher at Huazhong University of Science and Technology, China. 

Her research interests include robust control, mechatronic systems modelling and simulation, and deep learning applications.
\end{IEEEbiography}
\begin{IEEEbiography}[{\includegraphics[width=1in,height=1.25in,clip,keepaspectratio]{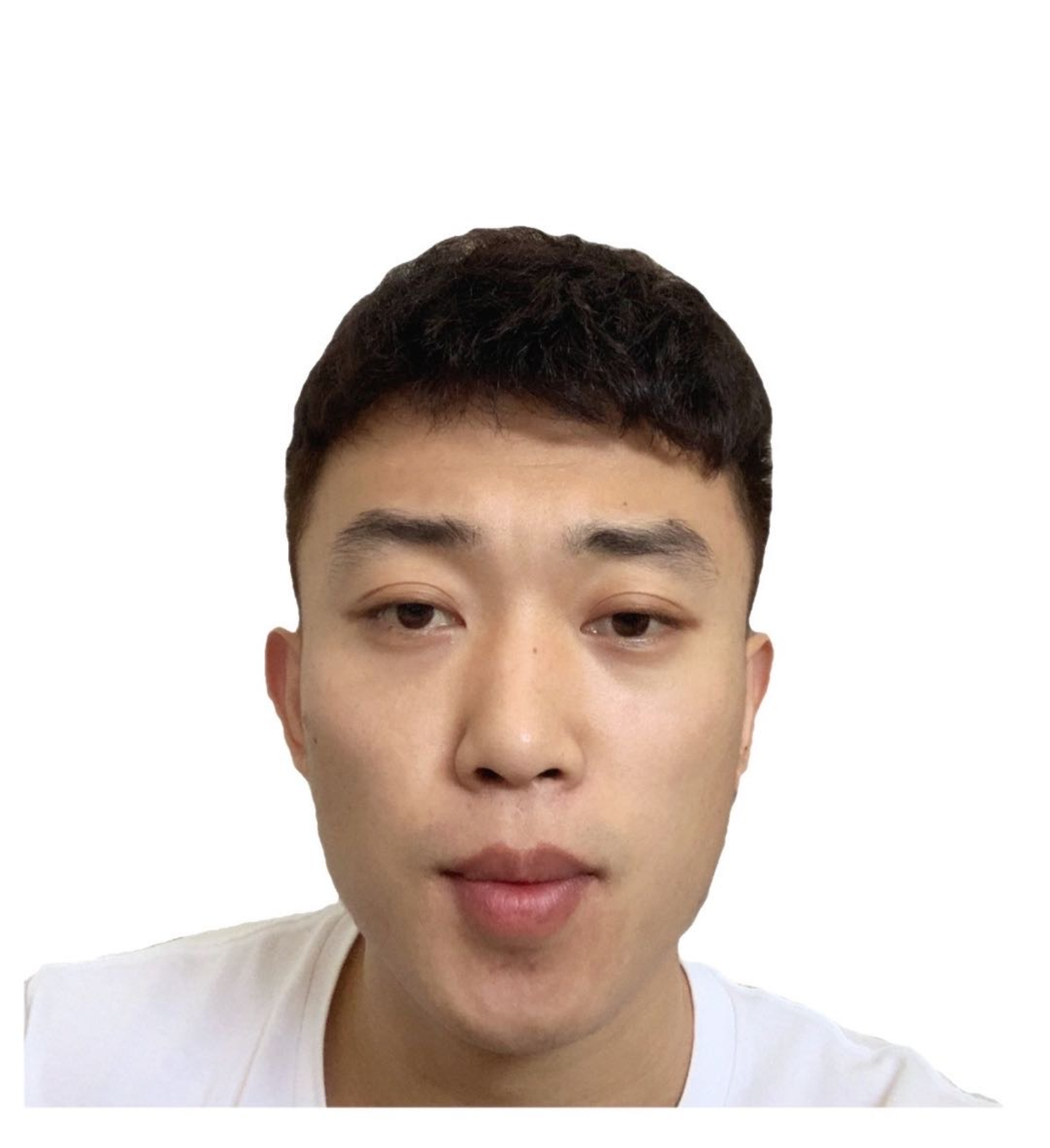}}]{Guijun Ma} received his Bachelor's degree from Huazhong University of Science and Technology, Wuhan, China, in 2017. He is currently working toward the Ph.D. degree in mechanical engineering in School of Mechanical Science and Engineering, Huazhong University of Science and Technology, Wuhan, China. 

His research interests include machine fault diagnosis and remaining useful life prediction. 
\end{IEEEbiography}
%
\begin{IEEEbiography}[{\includegraphics[width=1in,height=1.25in,clip,keepaspectratio]{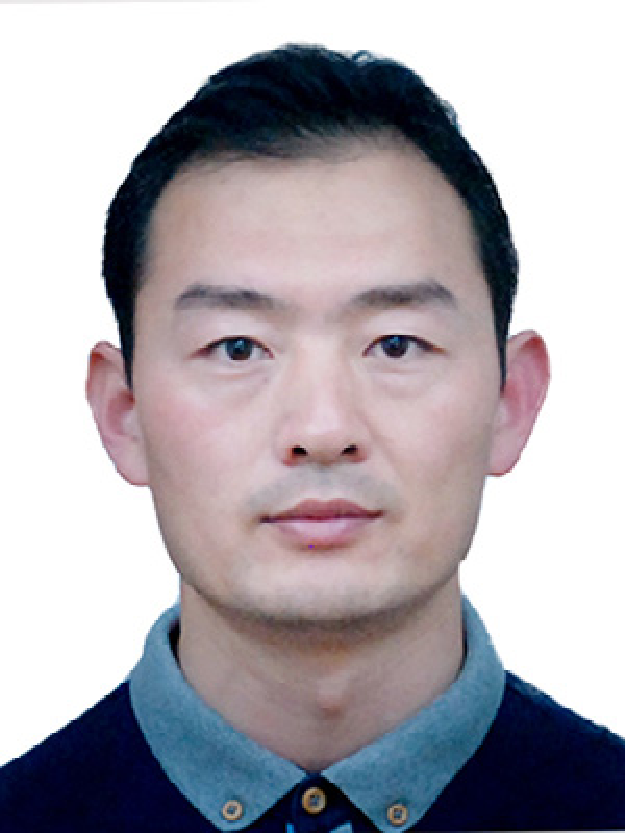}}]{Yong Zhang} received the B.Sc. degree in mathematics from Jiangsu Normal University, Xuzhou, China, in 2001, and the M.Sc. degree in applied mathematics from Three Gorges University, Yichang, China, in 2007. and the Ph.D. degree in control theory and control engineering from Huazhong University of Science and Technology, Wuhan, China, in 2010. From 2014 to 2015, he was a Visiting Scholar with the Department of Information Systems and Computing, Brunel University London, Uxbridge, U.K. He is currently an Associate Professor with the School of In- formation Science and Engineering, Wuhan University of Science and Technology, Wuhan, China. He has authored over 10 papers in refereed international journals. His cur- rent research interests include remaining useful life prediction of key equipment, fault diagnosis and fault tolerant control of networked systems. Dr. Zhang is a very active Reviewer for many international journals.
\end{IEEEbiography}
%
%
\begin{IEEEbiography}[{\includegraphics[width=1in,height=1.25in,clip,keepaspectratio]{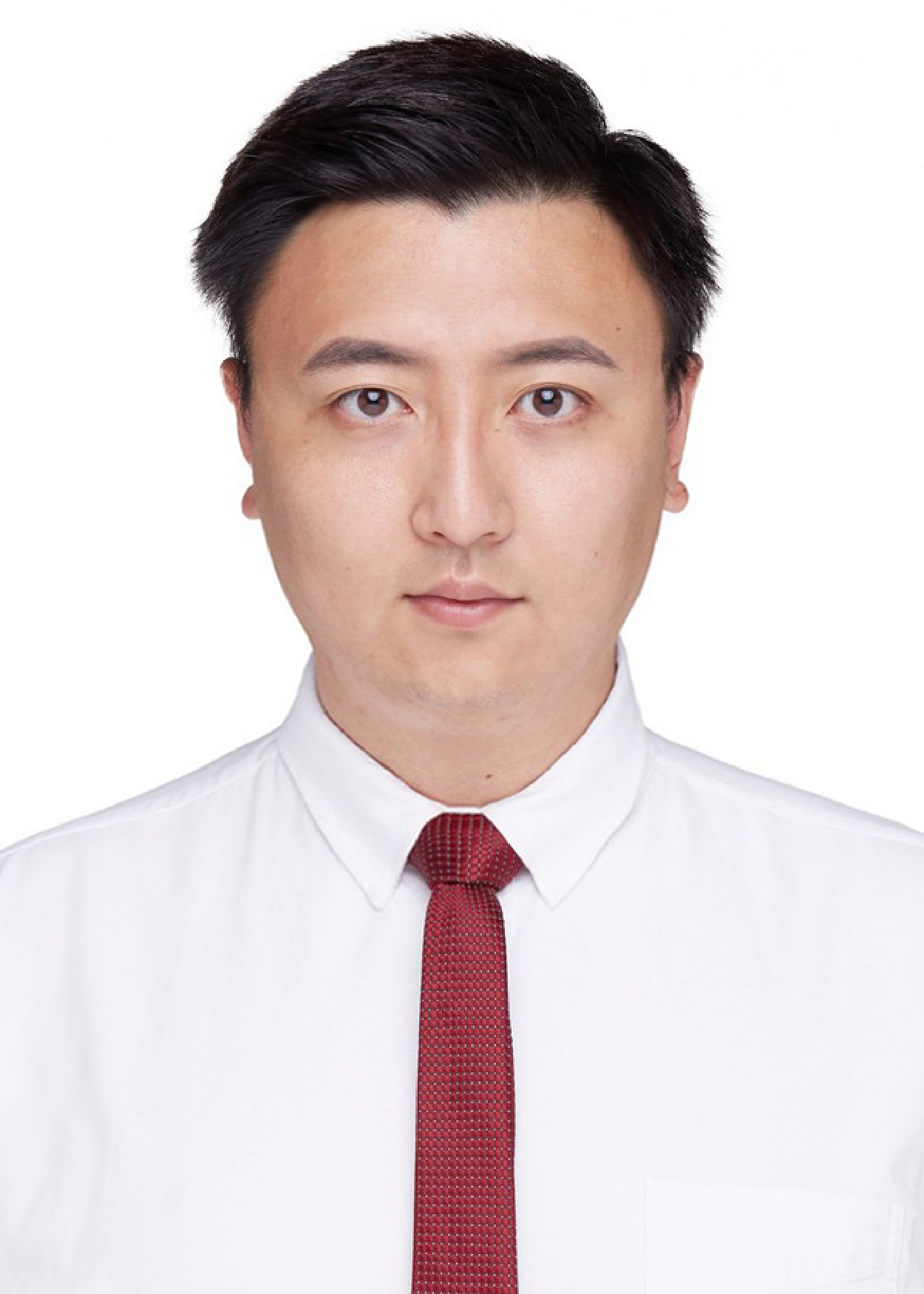}}]{Mingyang Sun (M'16)} received the Ph.D. degree from the Department of Electrical and Electronic Engineering in Imperial College London, U.K., in 2017. From 2017 to 2019, he was a Research Associate and a DSI Affiliate Fellow at Imperial College London.
 
He is now a Professor (Tenure-Track) under the Hundred Talents Program at Zhejiang University. Also, he is a Visiting Researcher at Imperial College London, UK. His research interests include Artificial Intelligence in Energy Systems and Cyber-Physical Energy System Security and Control.
\end{IEEEbiography}
\begin{IEEEbiography}[{\includegraphics[width=1in,height=1.25in,clip,keepaspectratio]{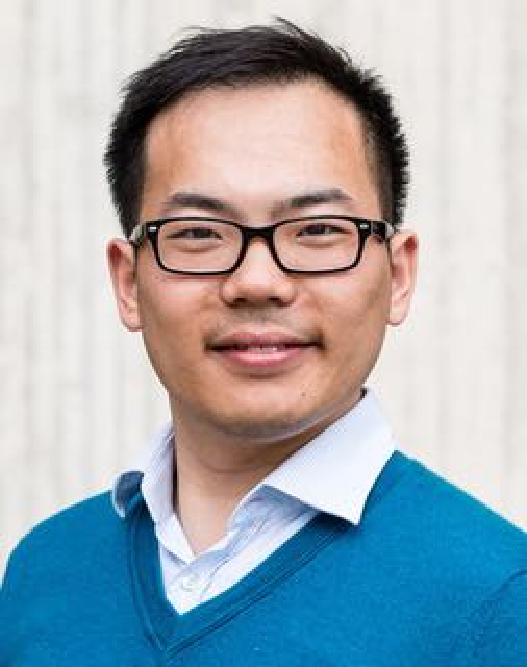}}]{Teng Fei (M'15)} received the BEng in Electrical Engineering from Beihang University, China, in 2009, and PhD degrees in Electrical Engineering from Imperial College London, U.K., in 2015. Currently, he is a Lecturer in the Department of Electrical and Electronic Engineering, Imperial College London, U.K. His research focuses on power system operation, cyber-physical system modelling, and data analytics.
\end{IEEEbiography}
%
\begin{IEEEbiography}[{\includegraphics[width=1in,height=1.25in,clip,keepaspectratio]{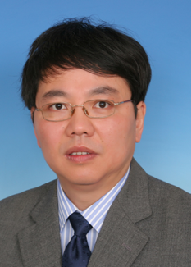}}]{Han Ding (M'97-SM'00)} received the Ph.D. degree in mechanical engineering from Huazhong University of Science and Technology (HUST), Wuhan, China, in 1989.

Supported by the Alexander von Humboldt Foundation, he was with the University of Stuttgart, Stuttgart, Germany, from 1993 to 1994. Since 1997, he has been a Professor at HUST, where he is currently the Director of the State Key Lab of Digital Manufacturing Equipment and Technology. He was a ``Cheung Kong" Chair Professor of Shanghai Jiao Tong University from 2001 to 2006. He was elected the member of Chinese Academy of Sciences in 2013. His research interests include robotics, multiaxis machining, and control engineering.

Dr. Ding served as an Associate Editor of IEEE TRANSACTIONS ON AUTOMATION SCIENCE AND ENGINEERING from 2004 to 2007. He is an Editor of IEEE TRANSACTIONS ON AUTOMATION SCIENCE AND ENGINEERING and a Senior Editor of IEEE ROBOTICS AND AUTOMATION LETTERS.
\end{IEEEbiography}
\begin{IEEEbiography}[{\includegraphics[width=1in,height=1.25in,clip,keepaspectratio]{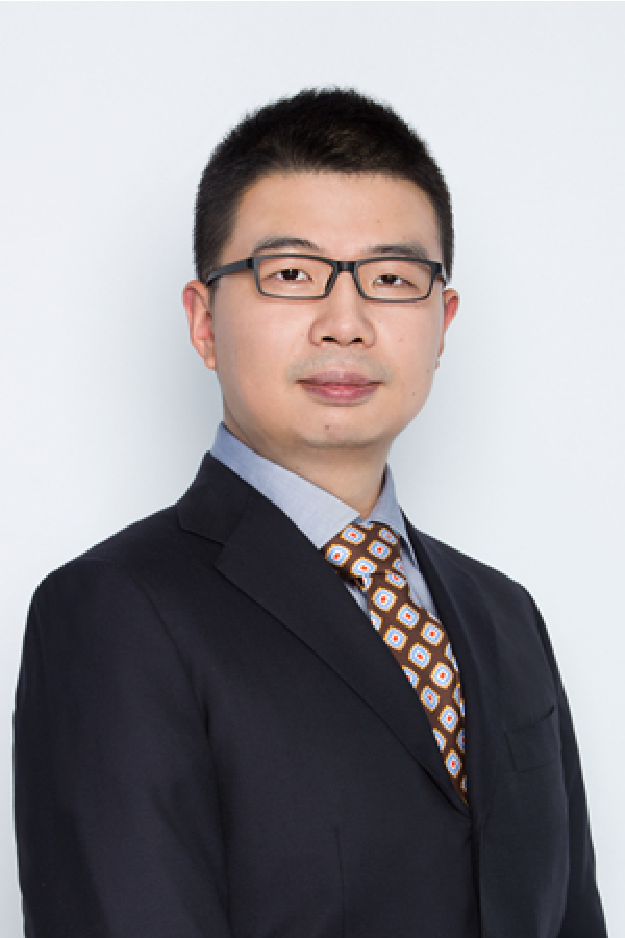}}]{Ye Yuan (M'13)} received the B.Eng. degree (Valedictorian) from the Department of Automation, Shanghai Jiao Tong University, Shanghai, China, in 2008, and the M.Phil. and Ph.D. degrees from the Department of Engineering, University of Cambridge, Cambridge, U.K., in 2009 and 2012, respectively, all in control theory. 

He has been a Full Professor with the School of Artificial Intelligence and Automation, Huazhong University of Science and Technology, Wuhan, China since 2016. Prior to this, he was a Postdoctoral Researcher at UC Berkeley, a Junior Research Fellow at Darwin College, University of Cambridge. His research interests include system identification and control with applications to cyber-physical systems. 

Dr. Yuan has received the China National Recruitment Program of 1000 Talented Young Scholars, the Dorothy Hodgkin Postgraduate Awards, Microsoft Research Ph.D. Scholarship, Best of the Best Paper Award at the IEEE Power and Energy Society General Meeting.
\end{IEEEbiography}


\end{document}